\RequirePackage{fix-cm}
\documentclass[smallextended]{svjour3}       
\smartqed  
\usepackage{graphicx}
\usepackage{subfigure}
\usepackage{hyperref}
\usepackage{amsmath,amssymb}
\usepackage[export]{adjustbox}
\usepackage{numprint}  

\usepackage{placeins} 

\DeclareMathOperator{\clip}{\texttt{clip}}
\DeclareMathOperator{\rnd}{\texttt{rnd()}}

\graphicspath{{./plots/}}
\DeclareGraphicsExtensions{.pdf,.png,.jpg}

\newcommand{\METHOD}{{KS(conf)\xspace}}

\usepackage{xspace}
\makeatletter
\DeclareRobustCommand\onedot{\futurelet\@let@token\@onedot}
\newcommand{\@onedot}{\ifx\@let@token.\else.\null\fi\xspace}

\newcommand{\eg}{{e.g}\onedot} 
\newcommand{\ie}{{i.e}\onedot}

\makeatother

\newcommand{\R}{\mathbb{R}}

\DeclareMathOperator{\argmax}{\operatorname{argmax}}

\begin{document}
\hyphenation{Conv-Net}
\hyphenation{NAS-NetA-large}

\title{\METHOD{}: A Light-Weight Test if a ConvNet Operates Outside of Its Specifications\thanks{This work was in parts funded by the European Research
Council under the European Union’s Seventh Framework Programme (FP7/2007-2013)/ERC grant agreement no 308036.}}

\titlerunning{A Light-Weight Test if a ConvNet Operates Outside of Its Specifications}        

\author{R\'emy Sun \and Christoph H. Lampert}

\institute{R. Sun, ENS Rennes, \email{remy.sun@ens-rennes.fr} \\
           C. H. Lampert, IST Austria, \email{chl@ist.ac.at}} 

\date{}

{
\def\makeheadbox{} 
\maketitle}

\begin{abstract}
Computer vision systems for automatic image categorization have become
accurate and reliable enough that they can run continuously for 
days or even years as components of real-world commercial applications.
A major open problem in this context, however, is quality control.
Good classification performance can only be expected if systems run 
under the specific conditions, in particular data distributions, 
that they were trained for. 
Surprisingly, none of the currently used deep network architectures 
has a built-in functionality that could detect if a network operates 
on data from a distribution that it was not trained for and 
potentially trigger a warning to the human users.

In this work, we describe \METHOD{}, a procedure for detecting 
such \emph{outside of the specifications} operation. Building on 
statistical insights, its main step is the applications of a
classical Kolmogorov-Smirnov test to the distribution of predicted 
confidence values.
We show by extensive experiments using ImageNet, AwA2 and DAVIS 
data on a variety of ConvNets architectures that \METHOD{} 
reliably detects out-of-specs situations. It furthermore has 
a number of properties that make it an excellent candidate 
for practical deployment: it is easy to implement, adds almost 
no overhead to the system, works with all networks, including 
pretrained ones, and requires no a priori knowledge about how 
the data distribution could change. 
\end{abstract}

\begin{figure}[t]
\centering
\begin{tabular}{c@{\qquad\qquad\qquad}ccc}
\fbox{\includegraphics[width=.15\textwidth]{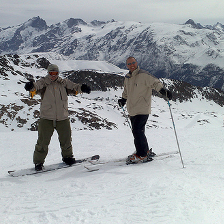}}&
\fbox{\includegraphics[width=.15\textwidth]{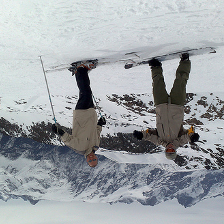}}&
\fbox{\includegraphics[width=.15\textwidth]{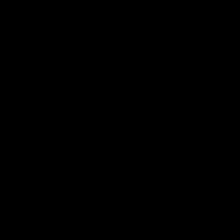}}&
\fbox{\includegraphics[width=.15\textwidth]{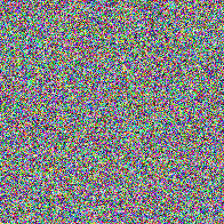}}
\\
$\downarrow$  & $\downarrow$ & $\downarrow$  & $\downarrow$ \\
\texttt{"ski"} & \texttt{"shovel"} & \texttt{"web site"} & \texttt{"tennis ball"}
\end{tabular}
\caption{Illustration of \emph{within specification} and \emph{outside of specifications} 
behavior of a ConvNet (here: VGG19, trained on ILSVRC2012). 
Left image: prediction on images that the network was trained to recognize.
Right images: prediction on images with different characteristics (rotated
180${}^\circ$, completely black, random noise). 
We observe: a standard multi-class network always predicts one of 
its predefined class labels, even if the current input is distorted, 
or even completely different, from what it was trained for.}
\label{fig:teaser}
\end{figure}

\section{Introduction}
With the emergence of deep convolutional networks (ConvNets), 
computer vision systems have become accurate and reliable enough 
to perform tasks of practical relevance autonomously and reliably
over long periods of time.
This has opened opportunities for the deployment of automated 
image recognition system in many commercial settings, such as
in video surveillance, self-driving vehicles, or social media.  

A major concern in our society about automatic decision systems 
is their reliability: if decisions are made by a trained classifier 
instead of a person, how can we be sure that the system works 
reliably now, and that it will continue to do so in the future? 
Formal verification techniques, as they are common for other 
safety-critical software components, such as device drivers, are 
still in their infancy for machine learning. Instead, quality 
control for trained system typically relies on extensive testing, 
making use of data that 1) was not used during training, and 2) 
reflects the expected situation at prediction time. 
If a system works well on a sufficiently large amount of data 
fulfilling 1) and 2), practical experience as well as machine 
learning theory tell us that it will also work well in the future. 
We call this \emph{operating within the specifications (within-specs)}.

In practice, problems emerge when a chance exists that the data 
distribution at prediction time differs from our expectations 
at training time, \ie when condition 2) is violated. 
Such \emph{operating outside of the specifications (out-of-specs)} 
can happen for a variety of reasons, ranging from user errors,
through changing environmental conditions, to problems with the 
camera setup, and even deliberate sabotage. 
Standard performance guarantees do not hold anymore in the 
out-of-specs situations, and the prediction quality often 
drop substantially.

While this phenomenon is well known in the research community, \eg under 
the name of \emph{domain shift}, and to some extent it is well understood, 
it remains a major obstacle for the deployment of deep learning system 
in real-world applications. 
Surprising as it is, today's most successful classification 
architectures, multi-class ConvNets with softmax activation, are not 
able to tell if they operate inside or outside the specifications. 
They will, for any input, predict one of the class labels they were 
trained for, no matter if the external situation matches the training 
conditions or not. 
Figure~\ref{fig:teaser} shows an example: a VGG19 ConvNet, trained on 
ImageNet ILSVRC2012, correctly predicts the label \texttt{ski} for the
left-most image. If that image is rotated by 180 degrees, though, 
\eg because the camera is mounted upside down, the network classifies 
it as \texttt{shovel}. As even more extreme examples, a completely 
black image is classified as \texttt{web site}, and a pattern of 
random noise as \texttt{tennis ball}.

One can easily imagine many ways for trying to solve this problem, 
for example, by training the network differently or changing its
architecture. 
However, any method that requires such changes to the training 
stage strongly reduces its real-world applicability. This is, because 
training ConvNets requires expert knowledge and large computational 
resources, so in practice most users rely on pretrained networks, 
which they use as black boxes, or fine-tune at best. Only the inputs 
can be influenced and the outputs observed. Sometimes, it might not 
even be possible to influence the inputs, \eg in embedded devices, 
such as smart cameras, or in security related appliances that prevent 
such manipulations.

It is highly desirable to have a test that can reliably tell when a 
ConvNet operates out-of-specs, \eg to send a warning to a human user, 
but that does not require modifications to any
ConvNet internals.
Our main contribution in this work is such a test, which is light-weight 
and theoretically well-founded.

We start in Section~\ref{sec:method} by formulating first principles 
that any test for out-of-specs operation should have. 
We then describe our proposed method, \METHOD{}, which has at its
core a classical Kolmogorov-Smirnov test that is applied to the 
confidences of network predictions.
\METHOD{} fulfills all required criteria and allows for simple 
and efficient implementation using standard components. 

In an extensive experimental evaluation we demonstrate the power of 
the proposed procedure and show its advantages over alternative 
approaches.
Over the course of these experiments, we obtained a number of 
insights into the diverse ways by which modern ConvNets reacts to 
change of their inputs. 
We find these of independent interest, so we report on a selection 
of them. 

\section{Testing for Out-of-Specs Operation}\label{sec:method}
Any test for out-of-specs operation should be able to determine 
if the conditions under which a classifier currently operates 
differs from the conditions it was designed for. 
Assuming a fully automatic system, the only difference that 
can occur is a change in the input data distribution between 
training/validation and prediction time. 
Furthermore, since no ground truth labels are available at 
prediction time, only changes in the marginal (\ie input image) 
distribution will be detectable.
These considerations lead to a canonical blueprint for 
identifying out-of-specs behavior: perform a statistical 
test if the data observed at prediction time and data from 
known within-specs operation originate from the same 
underlying data distribution. 

Unfortunately, statistical tests tend to suffer heavily from 
the \emph{curse of dimensionality}, so reliable tests between 
sets of ConvNet inputs (\eg high-dimensional images) is 
intractable.
Therefore, and because ultimately we are interested in the network 
predictions anyway, we propose to work with the distribution of 
network outputs.
For a multi-class classifier, the outputs are multi-dimensional, 
and modeling their distribution might still not be possible. 
Therefore, we suggest to use only the real-valued confidence value 
of the predicted label. 
These values are one-dimensional, available in most practical 
applications, and as our experiments show, they provide a 
strong proxy for detecting changes in the distribution of inputs. 
Because evaluating the network is a deterministic function, the 
output-based test cannot introduce spurious false positives: if 
reference and prediction distribution of the inputs are identical, 
then the distribution of outputs will be identical as well. 
It would in principle be possible that an output-based test 
overlooks relevant differences between the input distributions. 
Our experiments will show, however, that for a well-designed test
this does not seem to be the case. 

In addition to the probabilistic aspects, a test that aims at 
practical applications needs to have several additional features. 
We propose the following \emph{necessary conditions} that 
a test must satisfy:
\begin{itemize}
\item\textbf{universal.} The same test procedure should be 
applicable to different network architectures.
\item\textbf{pretrained-ready.} The test should be applicable 
to pretrained and finetuned networks and not 
require any specific steps during network training.
\item\textbf{black-box ready.} The test should not require 
knowledge of any ConvNet internals, such as the depth, 
activation functions, or weight matrices. 
\item\textbf{nonparametric.} The test should not require 
a priori knowledge \emph{how} the data distribution could change.
\end{itemize}

\subsection{Notation}
We assume an arbitrary fixed ConvNet, $f$, with $K$ softmax outputs, \ie 
for an input image $X$ we obtain an output vector $Y:=f(X)$ with $Y\in\R^K$. 
Assuming a standard multi-class setting with softmax activation function 
at the output layer, one has $0\leq Y[k]\leq 1$ for $k=1,\dots,K$ 
and $\sum_{k=1}^K Y[k]=1$. 
For any output $Y$, the predicted class label is $C:=\argmax_{k=1}^K Y[k]$, 
with ties broken arbitrarily. The confidence of this prediction is 
$Z := Y[C]$, \ie the $C$th entry of the output vector $Y$, or 
equivalently $Z=\max_{k=1}^K Y[k]$, \ie the largest entry of $Y$.

Treating $X$ as a random variable with underlying probability 
distribution $P_X$, the other quantities become random variables
as well, and we name the induced probability distributions 
$P_Y$, $P_C$, and $P_Z$ respectively.
We are interested in detecting if the distribution $P_Z$ at 
prediction time differs from the distribution when operating 
within the specifications. 

\subsection{\METHOD: Kolmogorov-Smirnov Test of Confidences}
We suggest a procedure for out-of-specs testing that we
call \METHOD{}, for \emph{Kolmogorov-Smirnov test of confidences}.
Its main component is the application of a Kolmogorov-Smirnov 
test~\cite{massey1951kolmogorov} to the distribution of confidence 
values. 
\METHOD{} consists of three main routines: \emph{calibration}, 
\emph{batch testing} and (optionally) \emph{filtering}. 

\paragraph{Calibration.}
In this step, we establish a reference distribution for 
within-specs operation. It is meant to be run when the 
classifier system is installed at its destination, but 
a human expert is still available who can ensure that the
environment is within-specs for the duration of the calibration
phase. 

To characterize the within-specs regime, we use a set of 
validation images, $X^{\text{val}}_1,\dots,X^{\text{val}}_n$  
and their corresponding network confidence outputs, 
$Z^{\text{val}}_1,\dots,Z^{\text{val}}_n$. 
For simplicity we assume all confidence values to be distinct.
In practice, this can be enforced by perturbing the values 
by a small amount of random noise. 

The distribution $P_Z$ is one-dimensional with known range.
Therefore, one can, in principle, estimate its probability 
density function (pdf) from a reasonably sized set of samples,
\eg by dividing the support $[0,1]$ into regular bins and 
counting the fraction of samples falling into each of them. 
For ConvNet confidence scores this would be wasteful, though, 
because these scores are typically far from uniformly distributed.
To avoid a need for data-dependent binning, \METHOD{} starts
by a processing step. First, we estimate the \emph{inverse cumulative distribution function}, $F^{-1}$, 
which is possible without binning: 
we sort the confidence values such that we can assume the 
values $Z^{\text{val}}_1,\dots,Z^{\text{val}}_n$ in monotonically 
increasing order.
Then, for any $p\in[0,1]$, the estimated \emph{inv-cdf} 
value at $p$ is obtained by linear interpolation:

\begin{align}
\hat F^{-1}(p) = \frac{k}{n} + \frac{p-Z^{\text{val}}_k}{n(Z^{\text{val}}_{k+1}-Z^{\text{val}}_k)} &\quad\text{for $k\in\{0,\dots,n\}$ with $p\in [Z^{\text{val}}_{k}, Z^{\text{val}}_{k+1}]$,} 
\end{align}
with the convention $Z^{\text{val}}_0=0$ and $Z^{\text{val}}_{n+1}=1$. 

A particular property of $F^{-1}$ is, that for $P_Z$-distributed $Z$,
the values $F^{-1}(Z)$ are distributed uniformly in $[0,1]$. 
Thus, the quantities we ultimately work with for \METHOD{} are the 
uniformized confidence scores $Z':=\hat F^{-1}(Z)$. 
This transformation will prove useful for two reasons: in the \emph{batch 
testing} phase, we will not have to compare two arbitrary distributions 
to each other, but only the currently observed distribution with the 
uniform one.
In the \emph{filtering} stage, where we do actually have to 
perform binning, we can use easier and more efficient uniform bins.

\paragraph{Batch testing.}
The main step of \METHOD{} is \emph{batch testing}, which determines
at any time of the classifier runtime, if the system operates 
within-specs or out-of-specs.
This step happens at prediction time after the system has been 
activated to perform its actual task. 

Testing is performed on batches of images, $X'_{1},\dots,X'_{m}$,
that can but do not have to coincide with the image batches that 
are often used for efficient Conv\-Nets evaluation on parallel 
architectures, such as GPUs.
To the corresponding network confidences we apply the inverse cdf 
that was learned during calibration, resulting in values, 
$Z'_{1},\dots,Z'_{m}$ that, as above, we consider sorted.
We then compute the Kolmogorov-Smirnov test statistics
\begin{align}
KS &:= \max\big( \quad\max_{k=1,\dots,m}\big\{ Z'_{k} - \frac{k-1}{m} \big\}, \quad \max_{k=1,\dots,m}\big\{ \frac{k}{m} - Z'_k \big\} \quad\big),
\end{align}
which reflects the biggest absolute difference between the (empirical) 
cdf of the observed batch, and the linearly increasing reference cdf.
For a system that runs within-specs, $Z'$ will be close to uniformly 
distributed, and $KS$ can be expected to be small. It will not be 
exactly $0$, though, because of finite-sample effects and random 
fluctuations.
It is a particularly appealing property of the KS statistics, that 
its distribution can be derived and used to determine confidence 
thresholds~\cite{massey1951kolmogorov}: for any $\alpha\in[0,1]$ there 
is a threshold $\theta_{\alpha,m}$, such that if we consider the test 
outcome \emph{positive} for $KS > \theta_{\alpha,m}$, then the expected 
probability of a false positive test results is $\alpha$. 
The values $\theta_{\alpha,m}$ can be computed numerically~\cite{JSSv008i18} 
or approximated well (in the regime $m\ll n$ that we are mainly 
interested in) by 
{\small$\theta_{\alpha,m}\approx\sqrt{\frac{-0.5\log(\alpha/2)}{m}}$.}

The Kolmogorov-Smirnov test has several advantages over other tests. 
In particular, it is \emph{distribution-free}, \ie the 
thresholds $\theta_{\alpha,m}$ are the same regardless of what 
the distribution $P_{Z'}$ was. Also, it is invariant under 
reparameterization of the sample space, which in particular means 
that the KS statistics and the test outcome we compute in comparing
$Z'$ to the uniform distribution is identical to the one for 
comparing the original $Z$ to the within-specs distribution.

\paragraph{Filtering.}
A case of particular interest of out-of-specs operation is when 
the working distribution is almost the reference one, but a certain 
fraction of unexpected data occurs, \eg images of object classes 
that were not present at training time.  
In such cases, simply send a warning to a human operator might not
be sufficient, because the difference in distribution is subtle and
might not become clear simply from looking at the input data.
We suggest to help the operator by highlighting \emph{which} of 
the images in the batch are suspicious and the likely reason
to cause the alarm. 
Unfortunately, on the level of individual samples it is not 
possible to identify suspicious examples with certainty, 
except when the support of the reference distribution and 
the unexpected data are disjoint. 
Instead, we suggest a \emph{filtering} approach: if the out-of-specs 
test is possible, we accompany the warning it with a small number 
of example images from the batch that triggered the warning. 
The example images should be chosen in a way to contain as high 
a fraction of samples from the unexpected component as possible. 

In \METHOD, we propose to use a density ratio criterion: if the 
desired number of example images is $w$ and the batch size is $m$,
we split the interval $[0,1]$ uniformly into $\lceil \frac{m}{w}\rceil$ 
many bins. We count how many of the values $Z'_1,\dots,Z'_m$ falls
into each bin, and identify the bin with the highest count. 
Because the average number of samples per bin is at least $w$, 
the selected bin contains at least that many samples. To reduce 
that number to exactly $w$, we drop elements randomly from it. 

The procedure can be expected to produce subset with a ratio of
suspicious to expected images as high as the highest ratio of 
density between then. 
This is, because the reference distribution is uniform, so samples 
from the reference distribution should contribute equally to each 
bin. A bin that contains many more samples than the expected value
is therefore likely the result of the unexpected distribution being
present. Furthermore, from the number of samples in the most populated 
bin one can derive an estimate what fraction of unexpected samples 
to expect in the subset. 

\paragraph{Resource Requirements.}
The above description shows that \METHOD{} can be implemented in 
a straight-forward way and that it requires only standard components,
such as sorting and binning.
The largest resource requirements occur during \emph{calibration}, 
where the network has to be evaluated for $n$ input inputs and 
the resulting confidence values have to be sorted.
The \emph{calibration} step is performed only once and offline though,
before actually before activating the classification system, so
$O(n\log n)$ runtime is not a major problem, and even very large 
$n$ remains practical. 
A potential issue is the $O(n)$ storage requirements, if calibration 
is meant to run on very small devices or very large validation sets.
Luckily, there exist specific data structures that allow 
constructing approximate \emph{cdf}s of arbitrary precision 
in an incremental way from streaming way. For example, 
\emph{t-digest}s~\cite{dunning2014computing} uses adaptive 
bins to build in empirical cdf in a streaming fashion. 
The memory requirements can be make $O(1)$ by this, with 
well-understood trade-offs between the memory footprint 
and the quality of approximating the cdf. 

The \emph{batch testing} step runs during the standard operation 
of the classification system and therefore needs to be as 
efficient as possible. 
Implemented as described above, it requires applying the inverse
cdf function, which typically requires a binary search, sorting 
the list of $m$ confidence values and identifying the maximum out 
of $2m$ values. Consequently, the runtime complexity is $O(m\log n)$ 
and the memory requirement is $O(m)$. 
With only logarithmic overhead, the added computational cost is 
typically negligible compared to evaluating the ConvNet itself. 
For even more restricted settings, incremental variants of the 
Kolmogorov-Smirnov test have been developed, see \eg~\cite{dos2016fast}.

The \emph{filtering} step adds very little overhead. As the transformed
confidence values are already available and the bins are regularly spaces,
creating the bin counts and finding the bin with the largest number of 
entries requires $O(m)$ operations and $O(m)$ memory.

\subsection{Related work}\label{subsec:related}
The problem of differences between data distributions at
training and prediction time are well known~\cite{ben2010theory}. 
Nevertheless, none of the existing ConvNet architectures for 
image classification have the built-in functionality to identify 
when they operate outside of their specifications. 
Instead, one typically attempts to ensure that out-of-specs 
situations will not occur, \eg by training invariant 
representations~\cite{ganin15}. Such \emph{domain adaptation} 
methods, however, require samples from the data distribution 
at prediction time and specific modifications of the network 
training, so they are not be applicable for end users who rely 
on pretrained classifiers. 
A notable exception is~\cite{royer-cvpr2015}, which adapts adapt 
a classifier on-the-fly at prediction time, but is limited to 
changes of the class proportions.
One specific out-of-specs situation are new classes that occur in 
the input data. Specific systems to handle these have been suggested 
for \emph{incremental}~\cite{icarl} 
or \emph{open set learning}~\cite{Bendale_2015_CVPR,Jain2014}.
These methods also modify the training procedure, though, and work 
only for specific classifier architectures.

Conceptually most related to our work is the field of \emph{concept drift} 
in the time series community, where the goal is to identify whether the 
distribution of a data stream changes over time.
The majority of works in this area are not applicable to the situation 
we study, because they require labeled data at prediction time 
(\eg \cite{harel2014concept,wang2015concept}) or suffer from the 
curse of dimensionality (\eg \cite{kuncheva2014pca,sethi2016grid}).
As an exception,~\cite{5693384} looks at concept drift detection from 
classifier output probabilities, and even suggests the use of a 
Kolmogorov-Smirnov test.
The method operates the context of binary classifiers, though, without 
a clear way to generalize the results to multi-class classification 
with large label sets.

\section{Experiments}
As most detection tasks, the detection of out-of-specs behavior 
can be analyzed in terms of two core quantities: the \emph{false 
positive rate (FPR)} should be as low as possible, and the 
\emph{true positive rate (TPR)} should be as high as possible.
While both quantities are important in practice, they typically
play different roles. 
The TPR should be high, because it measures the ability of the
test to perform the task it was designed for: detecting out-of-specs
behavior. 
The FPR should be low, because otherwise the test will annoy the
users with false alarms, and this typically has the consequence 
that the test is completely switched off or its alarms ignored. 
A practical test should allow its FPR to be controlled and ideally 
adjusted on-the-fly to a user-specific preference level. 
Our experimental protocol reflects this setting: for a user-specified  
FPR we report the TPR in a variety of settings that will be explained 
below. 

For \METHOD, adjusting the FPR is straightforward, as it immediately 
corresponds to the $\alpha$ parameter. 
When relying on the closed-form expression for the threshold, $\alpha$ 
can be changed at any time without overhead.
When using tabulated thresholds, $\alpha$ can at least be changed 
within the set of precomputed values. Note that, in particular, 
the expensive calibration step never has to be rerun. 
Many other tests, in particular some of the baselines we describe 
later, do not have this property. To adjust their FPR, one has to 
repeat their calibration, which in particular requires getting 
access to new validation data or storing the original data.

\subsection{Experimental Protocol}
To analyze the behavior of \METHOD{} and other tests in different 
scenarios, we perform a large-scale study using five popular ConvNet 
architectures: ResNet50~\cite{he2016deep} and VGG19~\cite{Simonyan14c} 
are standards in the computer vision community; SqueezeNet~\cite{iandola2016squeezenet}
and MobileNet25~\cite{howard2017mobilenets} have much smaller computational and memory  
requirements, which makes them suitable, \eg, for mobile and embedded  
applications; NASNetAlarge~\cite{zoph2017learning} achieves state-of-the-art performance 
in the ImageNet challenges, but is quite large and slow. 
Technical details of the networks can be found in Table~\ref{tab:networks}.

\newcommand{\oom}{\multicolumn{1}{c|}{---}}
\begin{table}[t]\scriptsize
\centering\begin{tabular}{|r|r|r||r|r|r||r|} 
\hline
\multicolumn{1}{|c|}{~} &
\multicolumn{1}{c|}{ILSVRC} &
\multicolumn{1}{c||}{number of} &
\multicolumn{3}{c||}{evaluation speed: GPU} &
\multicolumn{1}{c|}{CPU} 
\\
\multicolumn{1}{|c|}{network name} &
\multicolumn{1}{c|}{top-5 error} &
\multicolumn{1}{c||}{param.s} &
\multicolumn{1}{c|}{$bs\!=\!1$} & \multicolumn{1}{c|}{$bs\!=\!10$} & \multicolumn{1}{c||}{$bs\!=\!100$} &
\multicolumn{1}{c|}{$bs\!=\!1$} 
\\\hline
MobileNet25~\cite{howard2017mobilenets} & 24.2\% & 0.48\,M & 3.3\,ms & 5.2\,ms & 34\,ms & 682\,ms 
\\
SqueezeNet~\cite{iandola2016squeezenet} & 21.4\% & 1.2\,M & 5.7\,ms & 10.1\,ms & 113\,ms & 2288\,ms 
\\
ResNet50~\cite{he2016deep} & 7.9\% &	26\,M & 12.2\,ms & 34.1\,ms & 293\,ms & \oom 
\\
VGG19~\cite{Simonyan14c} & 10.2\% & 144\,M	& 9.9\,ms & 53.5\,ms & 385\,ms & \oom 
\\
NASNetAlarge~\cite{zoph2017learning} & 3.9\% & 94\,M & 45.8\,ms&227.9\,ms&2107\,ms & \oom 
\\\hline\end{tabular}
\caption{Details of the ConvNets used for the experimental evaluation.
Evaluation time 
(excluding image preprocessing and network initialization) 
for different batch sizes ($bs)$ on powerful GPU (NVIDIA Tesla P100) or 
weak CPU (Raspberry Pi Zero) hardware. Missing entries are due to memory limitations.}
\label{tab:networks}
\end{table}

As the main data source we use the ImageNet ILSVRC 2012 
dataset~\cite{ILSVRC15}, which consists of natural images 
of 1000 object categories. 
The ConvNets are trained on the 1.2 million training images\footnote{We use 
the pretrained models from~\url{https://github.com/taehoonlee/tensornets}.}. 
We use the 50.000 validation images to characterize the within-specs 
behavior in the calibration phase and the 100.000 test images to 
simulate the situation at prediction time. 
We do not make use of ground truth labels of the validation and test 
part at any time, as those would not be available for an actually 
deployed system, either.

Our experimental evaluation has two parts. First, we establish how 
well \METHOD{} and the baselines respects the chosen false positive 
ratios, and how well they are able to detect changes of the input image
distribution. 
We also benchmark \METHOD{}'s ability to identify a set of suspicious 
images out of a stream of images that contain ordinary as well as 
unexpected images.
Second, we provide further insight by analyzing how different network 
architectures react to a variety of changes in the input distribution, 
\eg due to sensor modifications, based on how easy or hard it is for 
\METHOD{} to detect those. 

\subsection{Baselines}\label{subsec:baselines}
There is no established standard on how to test if the distribution 
of network outputs coincide between training and deployment time. 
However, one can imagine a variety of way, and we include some 
of them as baselines in our experiments. 

\paragraph{Mean-based tests.} It is a generally accepted fact in
the community that Conv\-Nets are very confident in their decisions
for data of the type they were trained on, but less confident on 
other data. This suggests a straight-forward test for out-of-specs 
behavior: for a batch of images, compute the average confidence 
and report a positive test if that value lies below a threshold.

We include several baselines based on this reasoning that differ 
in how they set the threshold. The main two constructions are: 
\begin{itemize}
\item\textbf{$z$-test.} We compute the mean, $\mu$, and variance, 
$\sigma^2$, of the confidence values on the validation set. 
Invoking the law of large numbers, the distribution of the average 
confidence over a within-specs batch of size $m$ is approximately
Gaussian with mean $\mu$ and variance $\sigma^2/m$. 
We set the threshold to identify the lower $\alpha$-quantile of 
that distribution.
\item\textbf{(nonparametric) mean test.}
To avoid the assumption of Gaussianity, we use a bootstrap-like 
strategy: from the validation set we sample a large number of 
batches and we compute the mean confidence for each of them. 
The threshold is set such that at most a fraction of $\alpha$ 
of the batches are flagged as positive. 
\end{itemize}
Furthermore, we include a $\log$-$z$ and $\log$-\emph{mean} test. 
These do the same as the $z$ and \emph{mean}-test above, but 
work with the logarithms of confidences (logits) instead of 
the confidence values themselves.

The above tests are asymmetric, in the sense that they will 
detect if the mean confidence becomes too low, but not if 
its becomes too high.
To be sure that this is not a major limitation, we also 
include symmetric versions of the above tests: we determine 
two thresholds, an upper and a lower one, allowing for 
$\alpha/2$ false positives on each side. 

\paragraph{Label-based test.}
Instead of using the distribution of confidence values, it could 
also be possible to detect out-of-specs behavior from the distribution 
of actually predicted labels. 
\begin{itemize}
\item \textbf{$\chi^2$ test.} During calibration, we 
compute the relative frequency of labels on the validation set. 
For any batch, we perform a $\chi^2$  goodness-of-fit test, whether the 
empirical distribution is likely to originate from the stored one and report a positive test 
if the $p$-value lies below the desired FPR.
\end{itemize}

\subsection{Results: false positive rates}
As discussed before, it is a crucial property of a test to 
have a controllable (and ideally arbitrarily low) false 
positive rate. 
We check this for \METHOD{} and the baselines by running all 
tests on many batches sampled randomly from the ILSVRC test set. 
The distribution of this data is within-specs by assumption.
Therefore, all positive tests are false positives, and we
obtain the FPR by computing the fraction of positive tests. 

\begin{figure}[t]
\centering
\begin{tabular}{ccc}
\includegraphics[width=.31\textwidth]{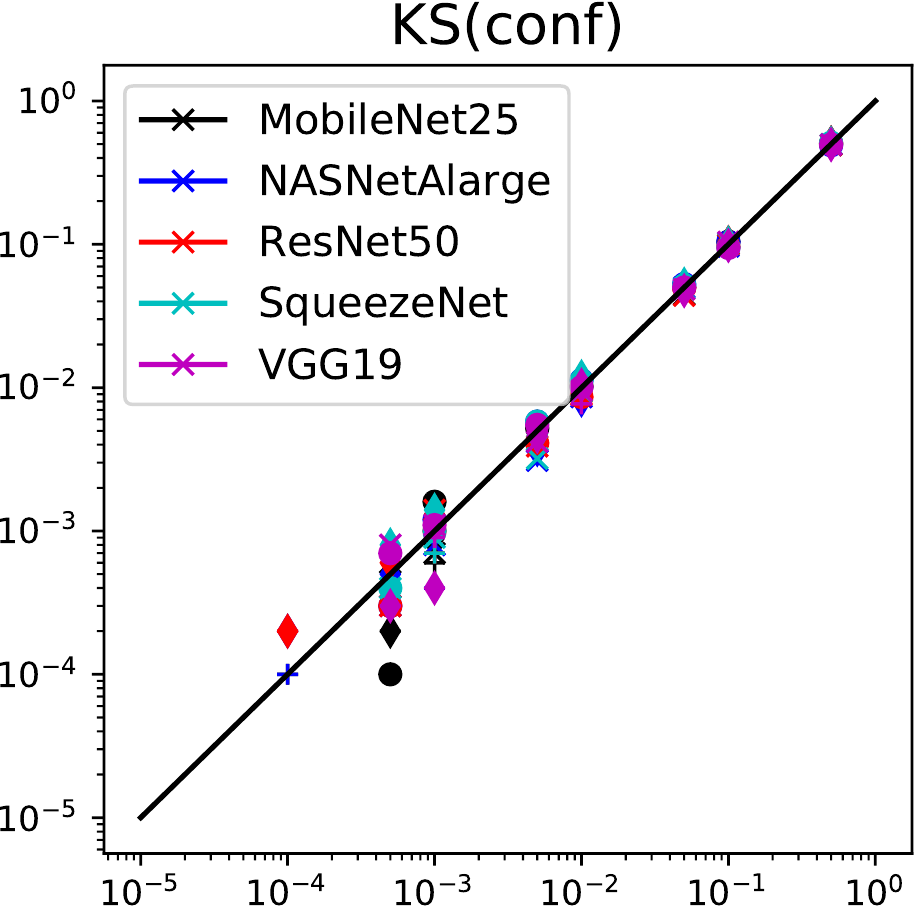} \\ 
\includegraphics[width=.31\textwidth]{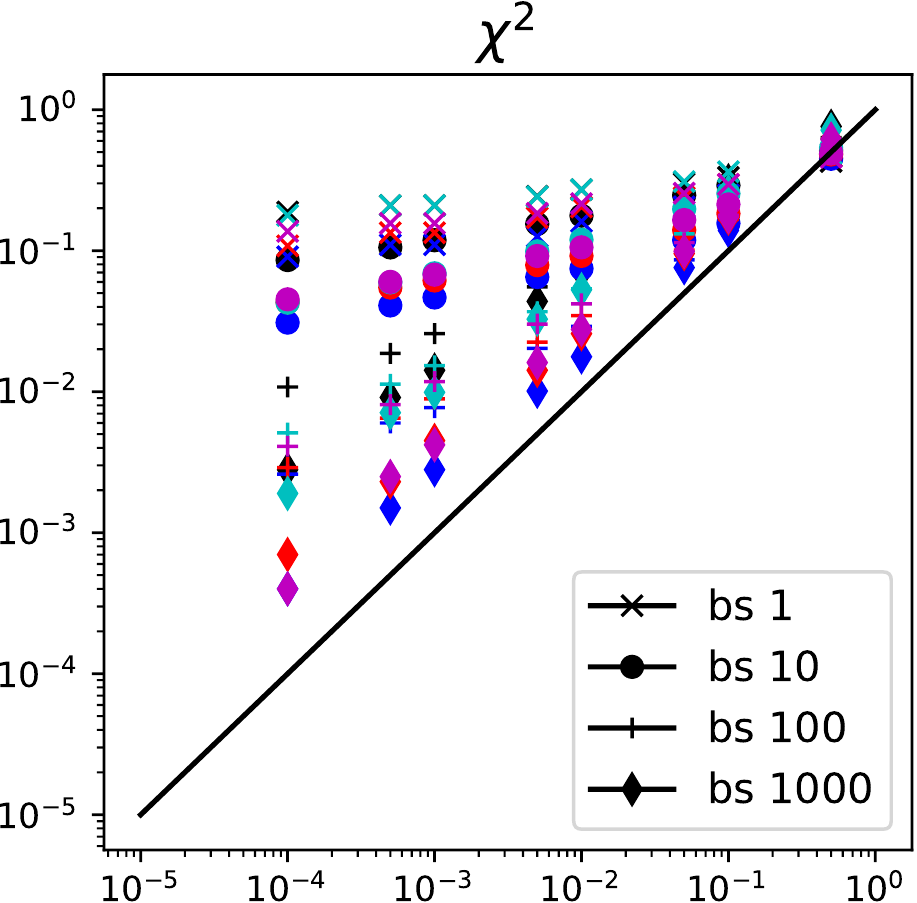} &
\includegraphics[width=.31\textwidth]{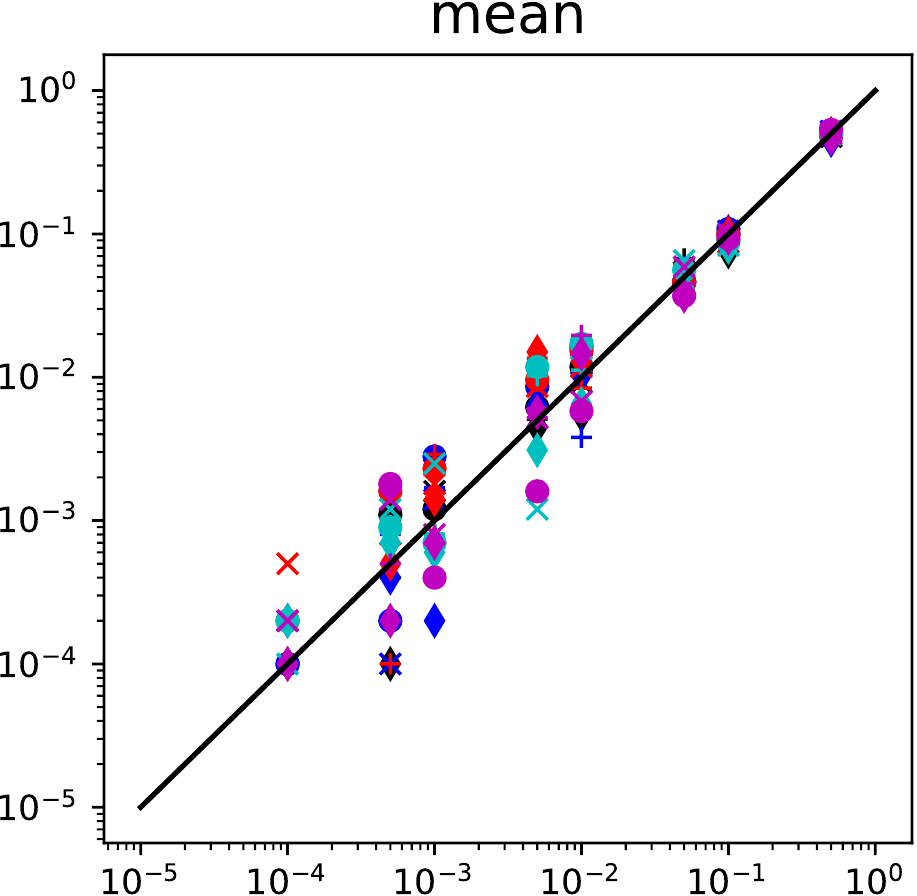} &
\includegraphics[width=.31\textwidth]{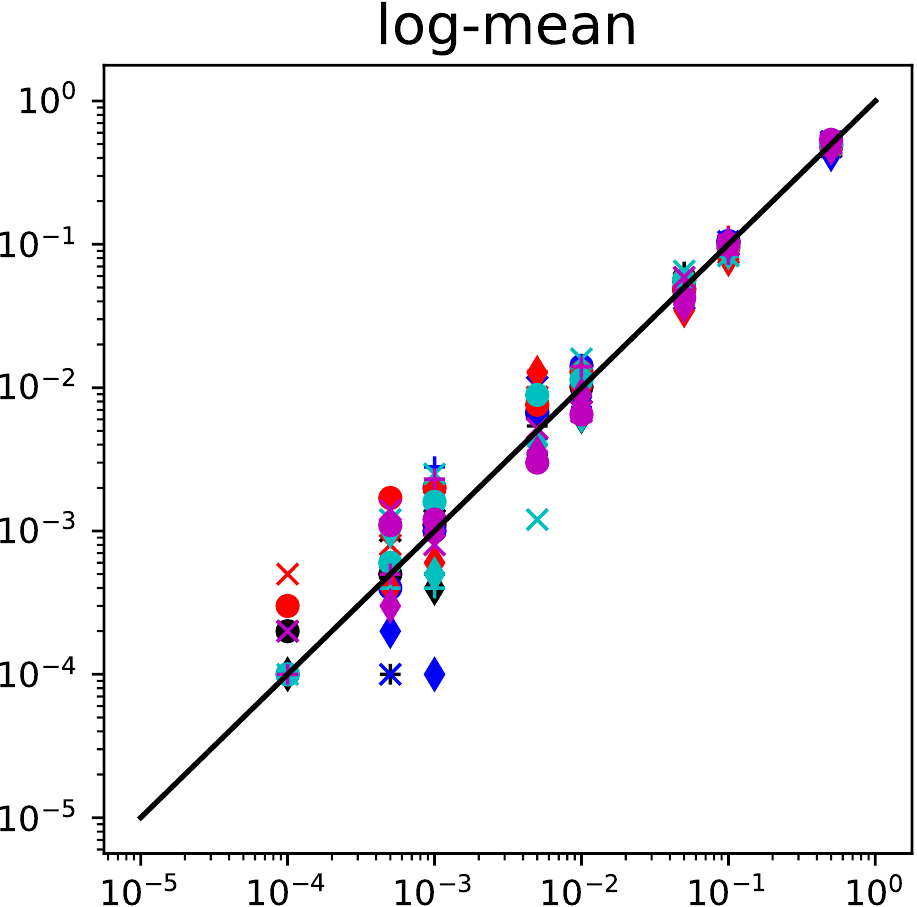} \\
\includegraphics[width=.31\textwidth]{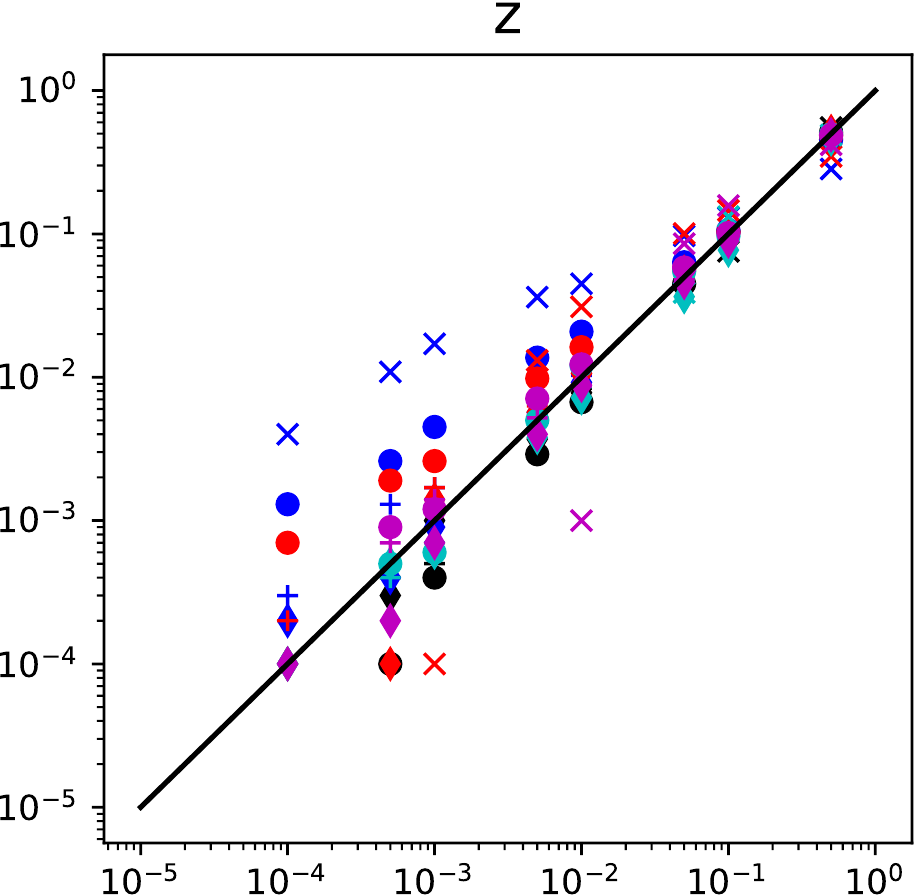} &
\includegraphics[width=.31\textwidth]{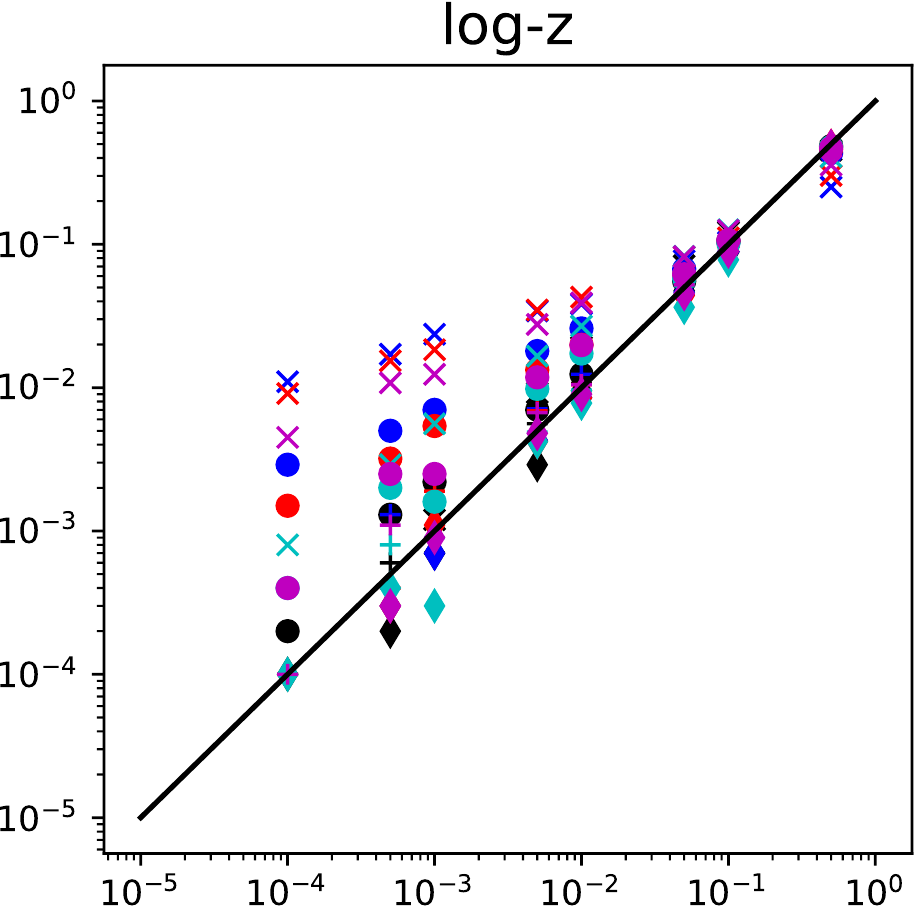} &
\includegraphics[width=.31\textwidth]{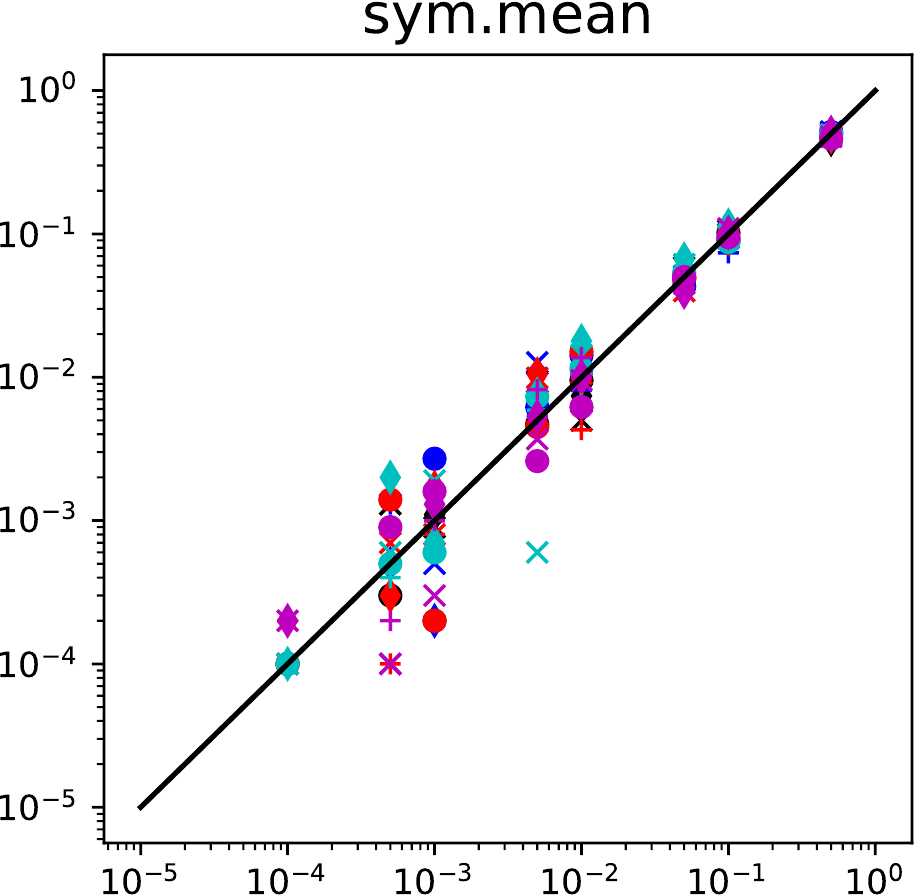} \\ 
\includegraphics[width=.31\textwidth]{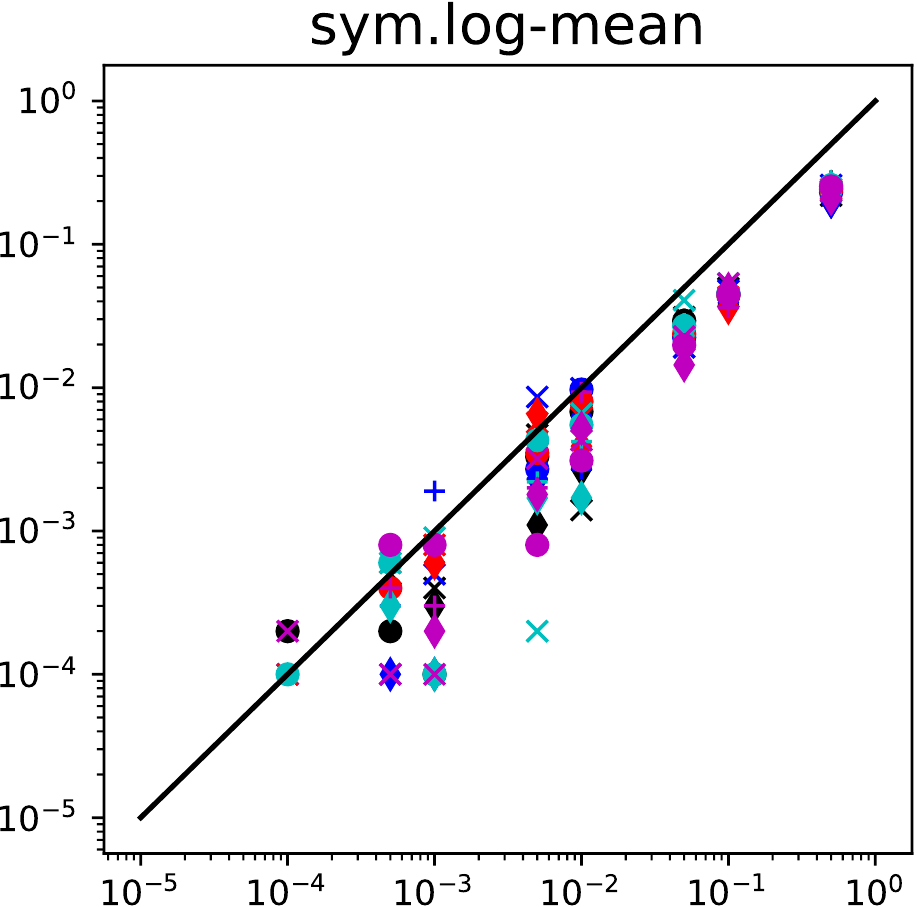} &
\includegraphics[width=.31\textwidth]{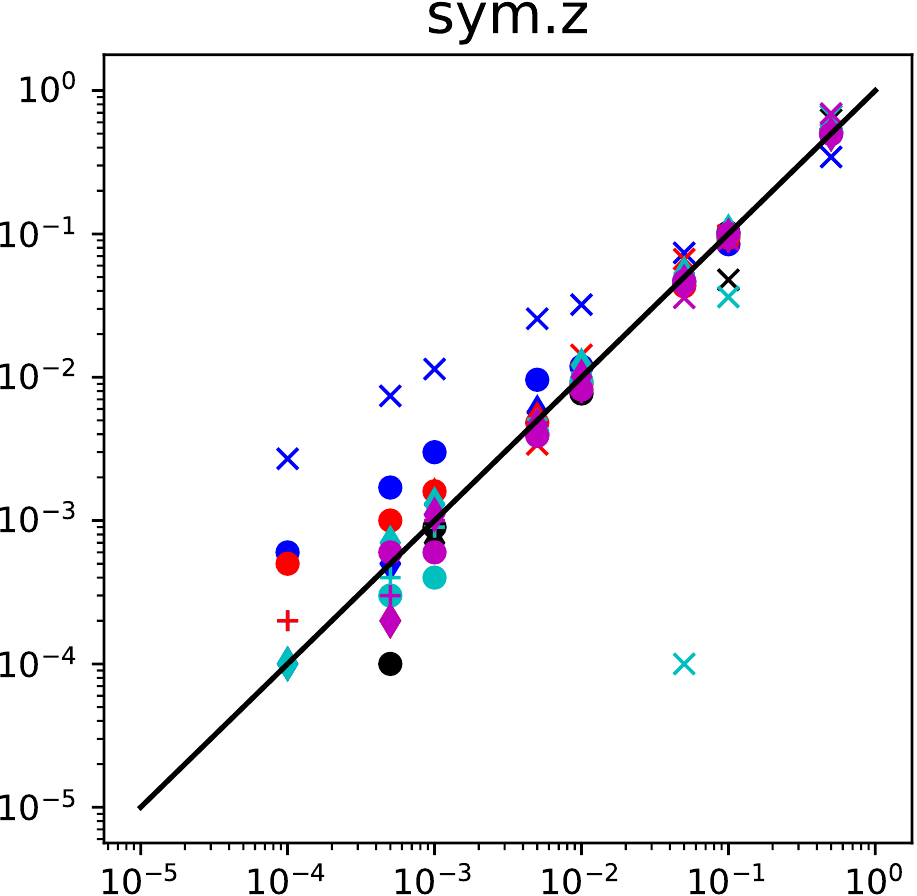}  &
\includegraphics[width=.31\textwidth]{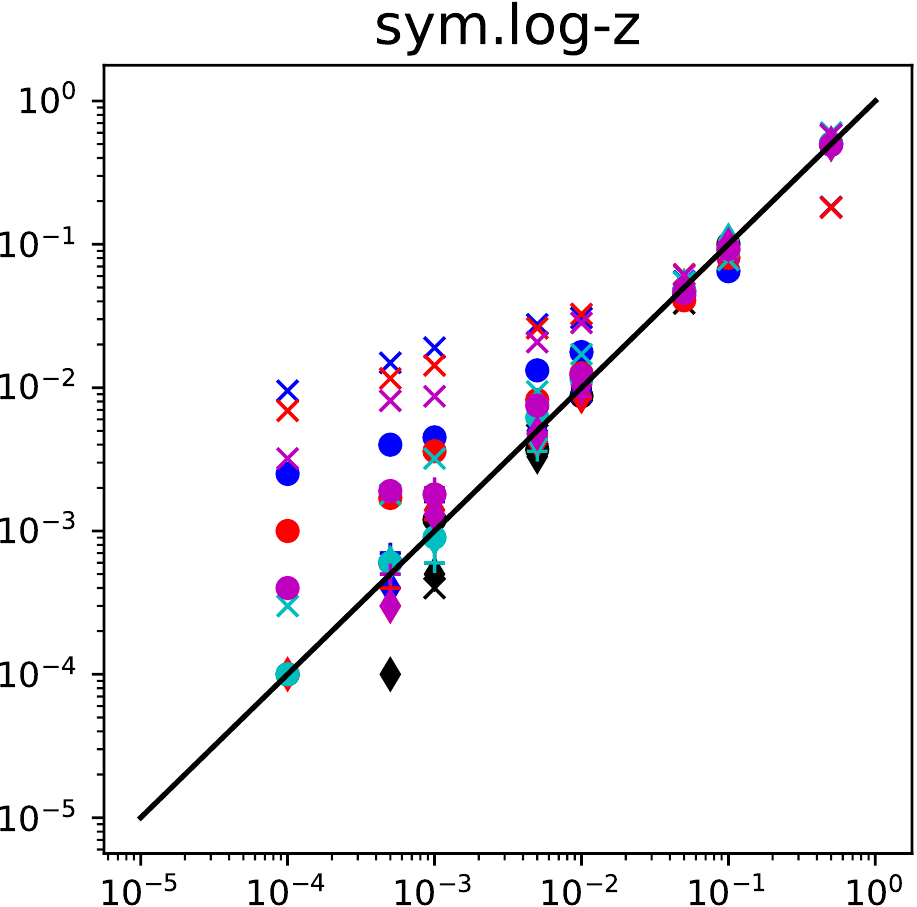}
\end{tabular}
\caption{False positive rates of \METHOD{} and baselines for 
different ConvNets and different batch sizes ($bs$). $x$-axis: target
FPR, $y$-axis: observed FPR. 
}\label{fig:FPR}
\end{figure}

The results are depicted in Figure~\ref{fig:FPR}, where we report 
the average FPR over 10.000 random batches.
One can see that \METHOD{}, \emph{mean} and \emph{log-mean} tests,
as well as their symmetric counterparts, respect the FPR well. 
For \METHOD{}, this is expected, as the underlying Kolmogorov-Smirnov 
test has well understood statistics and universal thresholds are available. 
For the other four tests, the thresholds are obtained from simulating 
the procedure of testing on within-specs data many time. This ensures 
that the FPR is respected, but it requires access to the validation 
data any time the FPR is meant to be adapted, and it can be very time 
consuming, especially if small FPRs are desired.

In contrast, $z$ and $\log$-$z$ tests and their symmetric variants 
often produce more false positives than intended, especially for 
small batch sizes. This is because their thresholds are computed 
based on an assumption of Gaussianity, which is a good approximation 
only for quite large batch sizes. 
Finally, the label-based $\chi^2$-test produces far too many false
positives. The likely reason for this is the high dimensionality of 
the data: a rule of thumb says that the $\chi^2$ test is reliable 
when each bin of the distribution has at least 5 expected entries.
This criterion is clearly violated in our situation, where the batch 
size is often even smaller than the number of bins. 
Consequently, we exclude the $\chi^2$-test from further experiments.
We keep the $z$ and $\log$-$z$ test in the list, but with the 
caveat that they should only be used with sufficiently large batch 
sizes.

In summary, of all methods, only \METHOD{} achieves the two desirable 
properties that the FPR is respected for all batch sizes, and that 
adjusting the thresholds is possible efficiently and without access 
to validation data.

\begin{table}[t]
\centering
\begin{tabular}{|l|r|r|r|r|r|r|r|r|r|r}
\hline
& 
\rotatebox{90}{\METHOD} & \rotatebox{90}{mean} & \rotatebox{90}{logmean} & \rotatebox{90}{$z$} & \rotatebox{90}{$\log$-$z$} & \rotatebox{90}{sym.mean} & \rotatebox{90}{sym.logmean} & \rotatebox{90}{sym.$z$} & \rotatebox{90}{sym.$\log$-$z$} 
\\\hline
AwA2-bat & 1.00 & 1.00 & 1.00 & 1.00 & 1.00 & 1.00 & 1.00 & 1.00 & 1.00 \\
AwA2-blue whale & 1.00 & 0.20 & 0.00 & 0.20 & 0.00 & 0.80 & 0.00 & 0.80 & 0.62   \\
AwA2-bobcat & 1.00 &  0.00 & 0.00 & 0.00 & 0.00 & 1.00 & 0.00 & 1.00 & 1.00 \\
AwA2-dolphin & 1.00 & 1.00 & 0.79 & 1.00 & 0.79 & 1.00 & 0.66 & 1.00 & 0.70  \\
AwA2-giraffe &  1.00  & 1.00 & 1.00 & 1.00 & 1.00 & 1.00 & 1.00 & 1.00 & 1.00  \\
AwA2-horse &   1.00 & 1.00 & 1.00 & 1.00 & 1.00 & 1.00 & 1.00 & 1.00 & 1.00  \\
AwA2-rat &     1.00  & 1.00 & 1.00 & 1.00 & 1.00 & 1.00 & 1.00 & 1.00 & 1.00  \\
AwA2-seal &   1.00  & 0.20 & 0.20 & 0.20 & 0.20 & 1.00 & 0.20 & 1.00 & 1.00 \\
AwA2-sheep &  1.00  & 0.01 & 0.00 & 0.00 & 0.00 & 0.65 & 0.00 & 0.63 & 0.82 \\
AwA2-walrus & 1.00  & 1.00 & 1.00 & 1.00 & 1.00 & 1.00 & 1.00 & 1.00 & 1.00   \\\hline
DAVIS & 1.00  & 1.00 & 1.00 & 1.00 & 1.00 & 1.00 & 1.00 & 1.00 & 1.00     \\\hline
\end{tabular}
\caption{True positive rate of \METHOD{} and baselines (averaged across $5$ ConvNets) 
under different out-of-specs conditions.}\label{tab:TPR}
\end{table}

\begin{figure}[t]
\includegraphics[width=\textwidth]{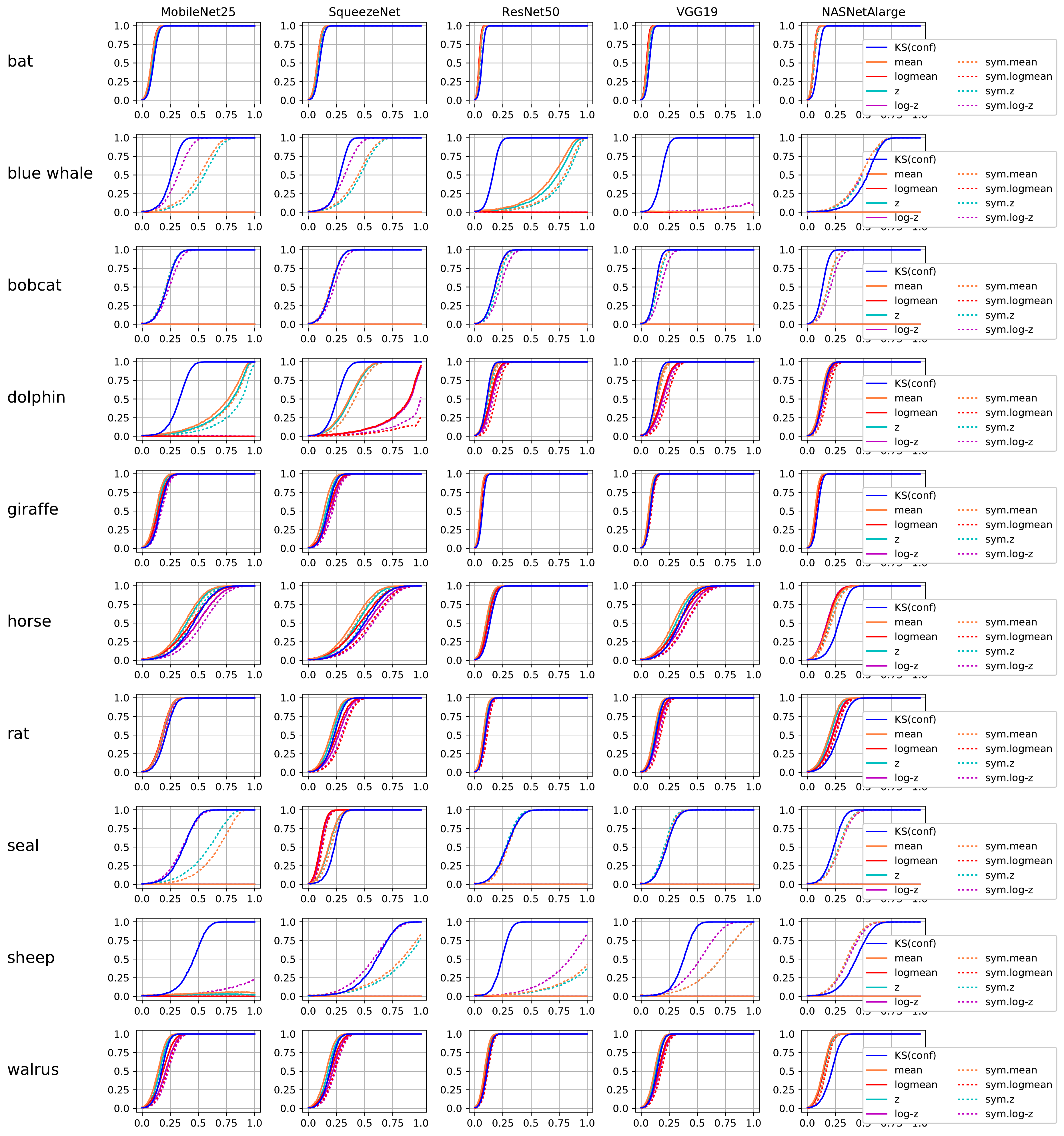}
\vskip\baselineskip
\includegraphics[width=\textwidth]{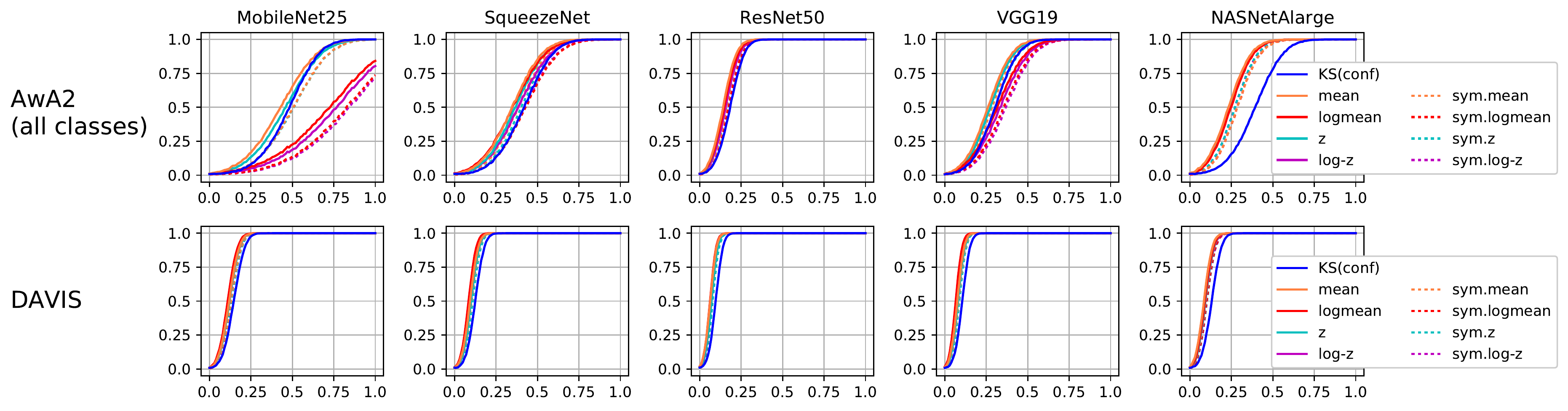}
\caption{Results of detecting out-of-specs behavior with different tests for different ConvNets and data sources. 
$x$-axis: proportions of unexpected (AwA2, DAVIS) vs. expected (ILSVRC) data in batch. $y$-axis: detection rate (TPR).}\label{fig:TPR}
\end{figure}

\begin{figure}[t]
\centering
\includegraphics[width=\textwidth]{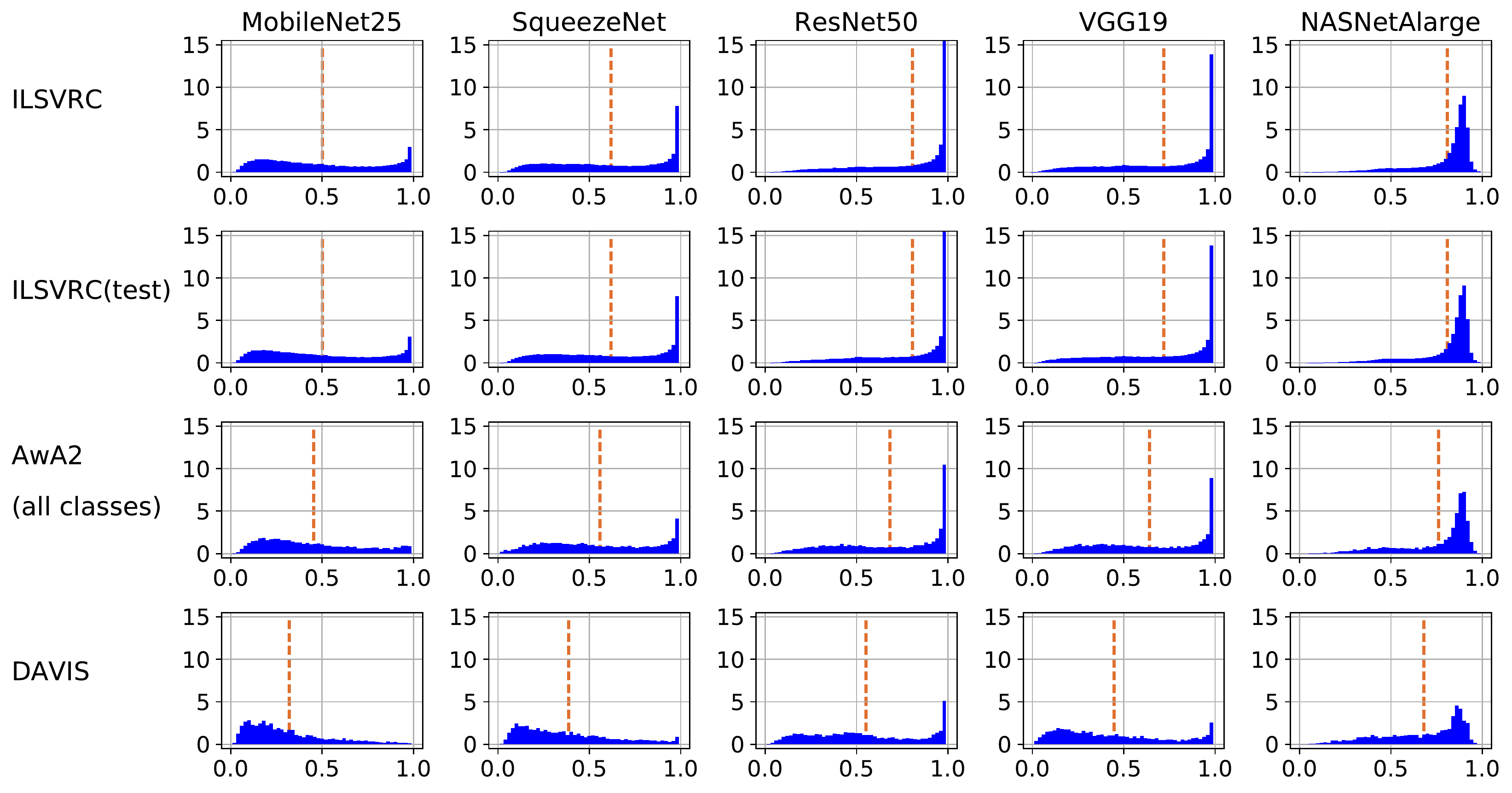}
\caption{Distribution of confidence scores for different ConvNets and data sources.}\label{fig:dist-AwA2}
\end{figure}
\begin{figure}
\centering
\includegraphics[width=\textwidth]{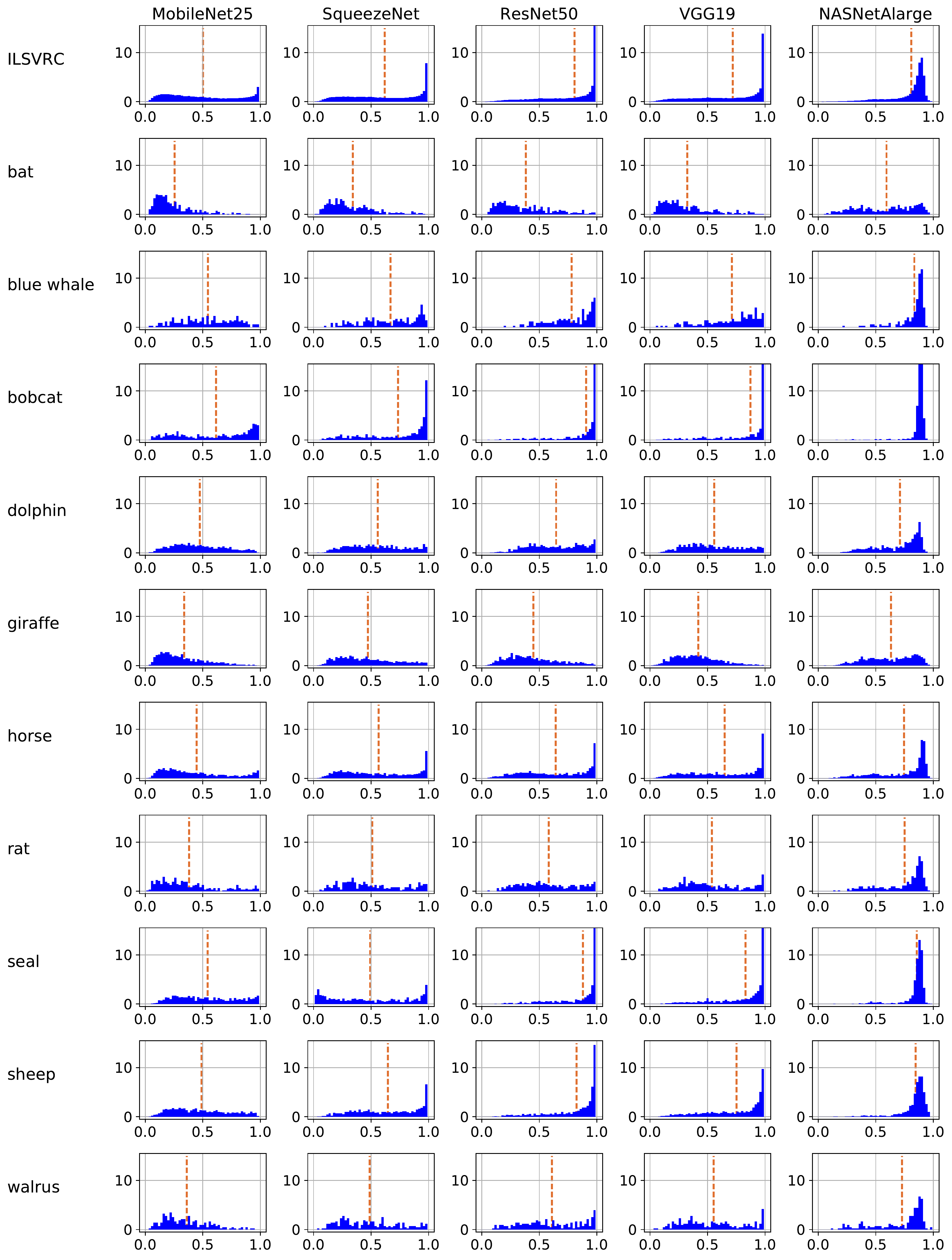}
\caption{Distribution of confidence scores for different ConvNets and data sources (cont).}\label{fig:dist-AwA2-cont}
\end{figure}

\subsection{Results: detection rate}\label{subsec:newclasse}
The true quality measure for any test is whether, at a fixed 
false positive rate, it is able to reliably detect changes 
in the input distribution.
To test this, we operate the ConvNets under out-of-specs conditions 
by evaluating them on image data that has different characteristics 
than ILSVRC. Specifically, we use the 7913 images from the 10 test 
classes of the \emph{Animals with Attributes 2 (AwA2) (proposed split)}~\cite{xian2017zero}.
These are natural images of similar appearance as ILSVRC, but from 
classes that are not present in the larger dataset.
Additionally, we also use the 3456 images from the 480p part of 
the DAVIS~\cite{Perazzi2016} dataset. 
The images are in fact video frames and therefore exhibit different 
characteristics than ILSVRC's still images, \eg motion blur. 

For randomly created batches, we test how often each of the tests 
is positive, and we report the average over 10.000 repeats. 
As the system is run completely out-of-specs in this scenario, 
all positive tests are correct, so the reported average directly 
corresponds to the TPR. %
Table~\ref{tab:TPR} shows the results for an FPR $\alpha=0.01$
and batch size $1000$ averaged across the five ConvNets. 
More detailed results can be found in Appendix~\ref{app:TPR}.

The main observation is that \METHOD{} is able to reliably detect 
the out-of-specs behavior for all data sources, but none of the
others methods are. All mean-based tests fails in at least some cases.

To shed more light on this effect, we performed more fine-grained 
experiments, where we create batches as mixtures of \emph{ILSVRC2012-test} 
and \emph{AwA2} images with different mixture proportions.
Figure~\ref{fig:TPR} shows the results as curves of TPR 
versus the mixture proportion. 
The results fall into three characteristic classes: 1) some sources, 
\eg \emph{bat}, are identified reliably by all tests for all ConvNets. 
2) other sources, \eg \emph{bobcat}, are identified reliably by some 
tests, but not at all by others. 3) for some sources, \eg \emph{blue whale}, 
tests show different sensitivities, \ie some tests work only for 
high mixture proportions.
Interestingly, the results differ substantially between networks.
For example, for the ResNet50, perfect detection is typically achieved 
at lower mixture proportions than for MobileNet25. For NASNetAlarge on 
at lower mixture proportions than for MobileNet25. For NASNetAlarge on 
\emph{blue whale} data, the symmetric mean and logmean tests work as 
least as well as \METHOD, but the same tests on the same data fail 
completely for VGG19. 

A possible source of explanation is Figure~\ref{fig:dist-AwA2}, 
where we plot the output distribution of different networks under
different conditions. 
The first two rows reflects within-specs behavior. To some extent, 
they confirm the folk wisdom that ConvNet scores are biased towards 
high values. However, one can also see a remarkable variability 
between different networks. For example, MobileNet25 has a rather 
flat distribution compared to, \eg, VGG19, and the distribution 
for NASNetAlarge peaks not at $1$ but rather at $0.9$.
The other rows of the figure show the distribution in the same 
out-of-specs situations at in Figure~\ref{fig:TPR},
allowing to understand the reasons for the described behavior.
For example, for \emph{bat}, the pattern is as one would expect 
from an unknown class: confidences are overall much lower, 
which explain why the difference in distribution is easy to 
detect for all tests. 
The distribution for \emph{blue whale} data differs much less 
from the within-specs situation, making it harder to detect. 
Finally, \emph{bobcat} shows truly unexpected behavior: the 
networks are overall more confident in their predictions, which 
explains why in particular the single-sided mean-based tests 
fails in this case. 

Overall, our experiments show that \METHOD{} works reliably in 
all experimental conditions we tested. This is in agreement with 
the expectations from theory, as the underlying Kolmogorov-Smirnov 
test is known to have asymptotic power 1, meaning that if given 
enough data, it will detect any possible difference between
distributions. 
In contrast to this, the mean-based tests show highly volatile behavior, 
which makes them unsuitable as a reliable out-of-specs test.

\begin{figure}[t]
\includegraphics[width=\textwidth]{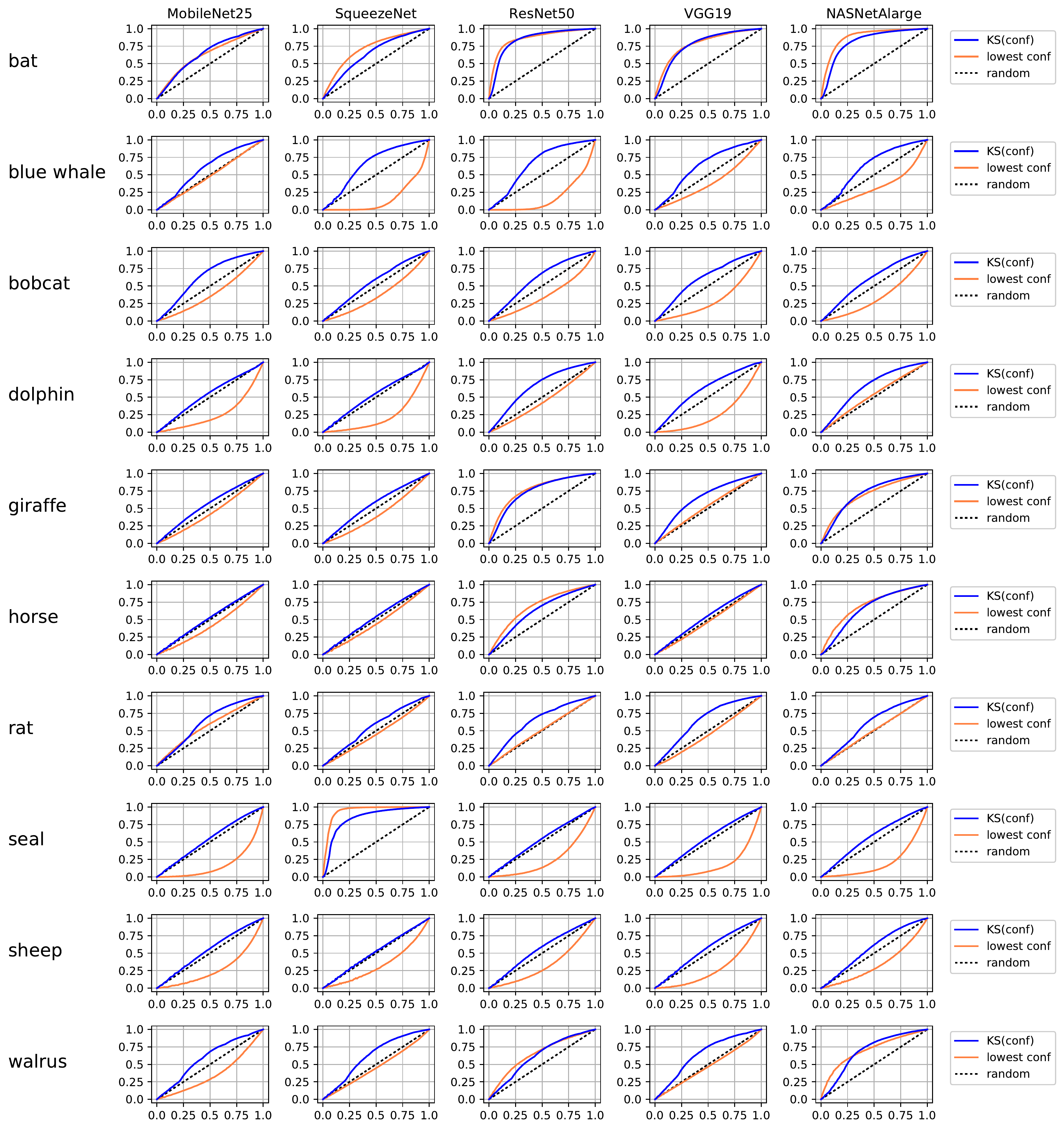}
\caption{Results of \emph{filtering}, \ie identifying which 
images in a batch likely corresponds to out-of-specs operation.
$x$-axis: fraction of unexpected (AwA2) versus expected (ILSVRC)
images in batch. $y$-axis: fraction of unexpected versus 
expected images in produced subset.}\label{fig:suspicious}
\end{figure}

\subsection{Results: filtering}\label{subsec:filtering}
To benchmark the ability of \METHOD{} to filter out images of
an unexpected distribution, we run the filtering routine in the 
setting of the previous section, \ie when the data at prediction 
time is a mixture of ILSVRC and AwA2 images. 
Each time \METHOD{} reports a positive test, we identify a subset 
of $10$ images\footnote{Other subset sizes did not yield 
substantially different results.} and measure how many of 
them were indeed samples from the unexpected classes.
There is no established baseline for this tasks, so we compare 
against the intuitive baseline of reporting as subset the $10$
images with lowest confidence. 

Figure~\ref{fig:suspicious} shows the fraction of outlier 
images in the identified subset for varying fractions 
of outlier images in the batch. 
A curve above the diagonal means that the test was successful 
in identifying suspicious images, a line below indicates that 
the method performs worse than a random selection.
One can see that \METHOD{} consistently creates better subsets
than random selection would. 
Selecting the images of lowest confidence instead sometimes 
works very well (\eg for \emph{bat}), but fails in other cases 
(\eg \emph{blue whale, bobcat}), where it performs worse than 
random sampling. As can be expected, the reason lies in the 
fact that ConvNet confidences do not always decrease during 
out-of-specs operation, as we had already observed in 
Figure~\ref{fig:dist-AwA2}.

\subsection{Results: camera system changes}
Another important reason why automatic imaging systems could 
operate out-of-specs is changes to the camera system.
We benchmark the performance of \METHOD{} to detect such, 
sometimes subtle, changes by running it on images that we obtain 
from the ILSVRC test set by applying characteristic manipulations.
Because of the results of the previous section, we only report 
on the results of \METHOD{} here. Our main goal is to gain 
insight into how different ConvNets react if the their inputs 
change due to external effect, such as incorrect camera installation, 
incorrect image exposure, or broken sensor pixels.

Specifically, we perform the following manipulations. Details of 
their implementation can be found in the corresponding sections.
\begin{itemize}
\item \textit{loss-of-focus}: we blur the image by filtering 
with a Gaussian kernel,
\item \textit{sensor noise}: we add Gaussian random noise to each pixel,
\item \textit{dead pixels}: we set a random subset of pixels to pure black or white,
\item \textit{wrong geometry}. we flip the image horizontally or vertically,
or rotate it by 90, 180 or 270 degrees
\item \textit{incorrect RGB/BGR color processing}: we swap the $B$ and $R$ color channel,
\item \textit{over- and under-exposure}: we scale all image intensities towards $0$ or $255$.
\end{itemize}

\begin{figure}[t]\centering
\begin{tabular}{cccc}
\includegraphics[width=.22\textwidth]{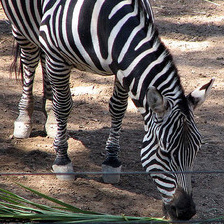}&
\includegraphics[width=.22\textwidth]{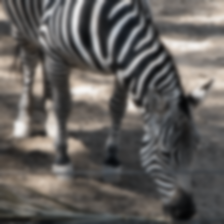}&
\includegraphics[width=.22\textwidth]{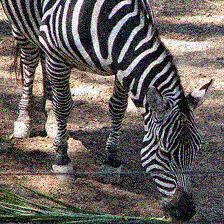}&
\includegraphics[width=.22\textwidth]{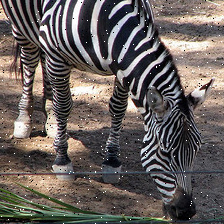}\\
original & loss-of-focus  & sensor noise & dead pixels\medskip\\
\includegraphics[width=.22\textwidth]{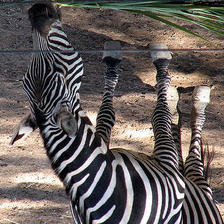}&
\includegraphics[width=.22\textwidth]{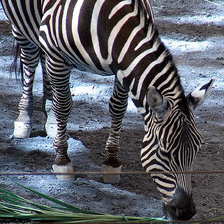}&
\includegraphics[width=.22\textwidth]{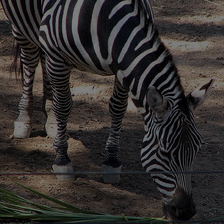}&
\includegraphics[width=.22\textwidth]{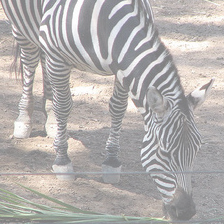}\\
wrong geometry & incorrect colors & under-exposure & over-exposure
\end{tabular}
\caption{Illustration of the effect of camera system changes on the input images.}\label{fig:synthetic}
\end{figure}

Figure~\ref{fig:synthetic} illustrates the operations. More detailed 
examples are provided in Appendix~\ref{app:synthetic}. 
In the following section, we discuss our main findings. A more in-depth 
analysis would likely find further noteworthy effects, though. 
Therefore, we have released the raw data and source code of the 
analysis for public use.\footnote{Code and data are available at \url{https://github.com/ISTAustria-CVML/KSconf}.}. 

\subsubsection{Loss of focus}
\begin{figure}\centering
\adjincludegraphics[height=.3\textwidth,trim={0 0 {.47\width} 0},clip]{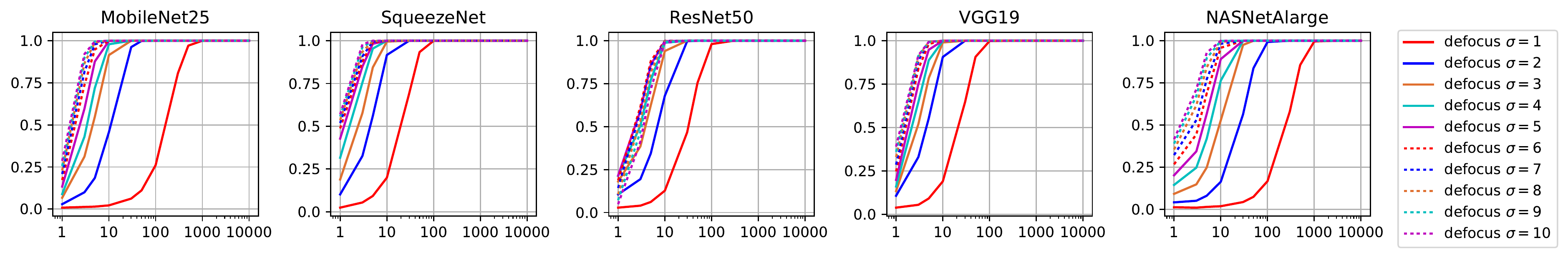}

\adjincludegraphics[height=.3\textwidth,trim={{.54\width} 0 0 0},clip]{batchsizes-blur}
\caption{Detection rates vs. \METHOD{} batch size for camera defocus (Gaussian blur) of different strengths.}\label{fig:batchsizes-blur}
\end{figure}

\begin{figure}\centering
\includegraphics[width=\textwidth]{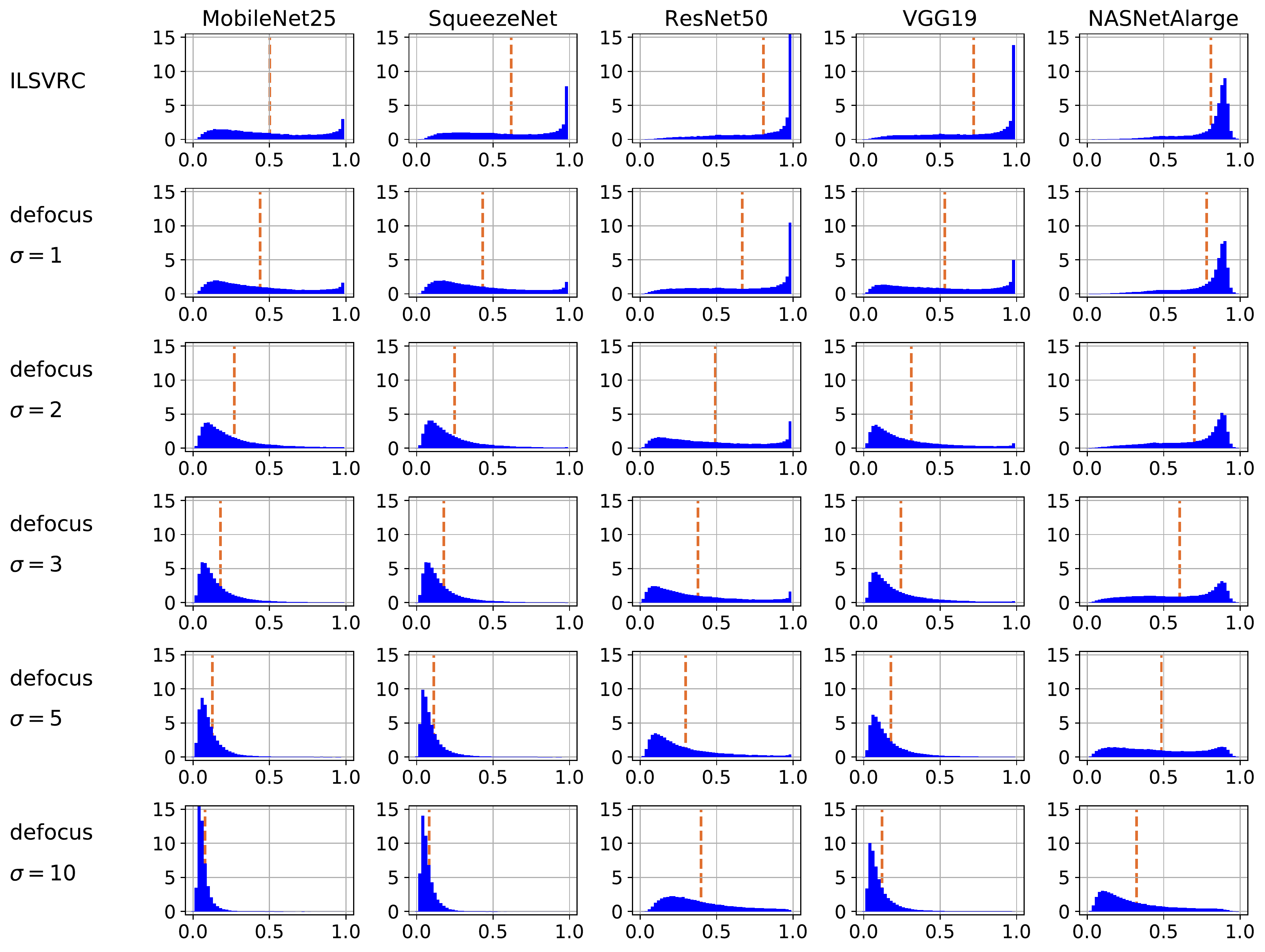}
\caption{Confidence scores for camera defocus (Gaussian blur) of different strengths. The dashed line marks the mean confidence.}\label{fig:distribution-blur}
\end{figure}

To analyze the effect of a loss of focus on the ConvNet outputs,
we create 100.000 perturbed test images by applying a Gaussian filter 
with variance $\sigma$, for each $\sigma\in\{1,2,\dots,10\}$. 
Note that we filter in horizontal and vertical directions, but not 
across color channels. 
$$\tilde X_t = X_t \ast g_{\sigma}\qquad \text{for $t=1,\dots,100.000$}$$
Figure~\ref{fig:batchsizes-blur} shows the detection rate 
achieved by \METHOD{} for different batch sizes.  
Figure~\ref{fig:distribution-blur} shows the resulting distribution
of confidence scores. 
One can see that already a rather weak loss of focus (\ie blur)
has a noticeable impact on the distribution of confidence scores. In particular,
for \emph{SqueezeNet}, \emph{ResNet50} and \emph{VGG19}, the strong peak of
confidence scores at or very close to $1$ is reduced. 
We observe that a loss of focus can reliably be detected from the 
confidence scores, and that this result seems rather stable across 
different ConvNet architectures. 

\clearpage

\clearpage
\subsubsection{Sensor Noise}
To analyze the effect of sensor noise, we create 100.000 perturbed 
test images by adding independent Gaussian noise with variance $\sigma$ 
in all color channels, for $\sigma\in\{5,10,15,20,30,50,100\}$:
$$\tilde X_t[h,w,c] = \clip_{0}^{255}\big(X_t[h,w,c] + \sigma\rnd\big) \qquad \text{for $t=1,\dots,100.000$,}$$
where $\rnd$ generates samples from a standard Gaussian distribution
and $\clip_\text{0}^\text{255}(\cdot)$ denotes the operation to clip 
a value to the interval $[0,255]$.

\begin{figure}
\centering
\adjincludegraphics[height=.3\textwidth,trim={0 0 {.47\width} 0},clip]{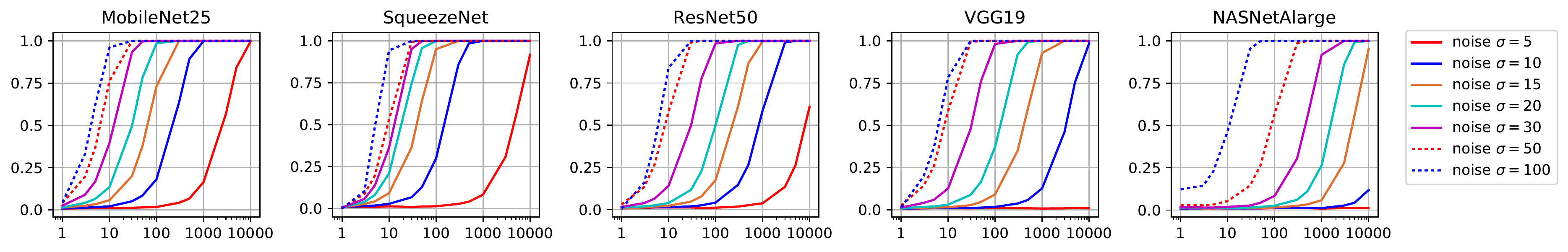}

\adjincludegraphics[height=.3\textwidth,trim={{.54\width} 0 0 0},clip]{batchsizes-noise}

\caption{Detection rates vs. \METHOD{} batch size for sensor noise (additive Gaussian noise) of different strengths}\label{fig:batchsizes-noise}
\end{figure}

\begin{figure}
\centering
\includegraphics[width=\textwidth]{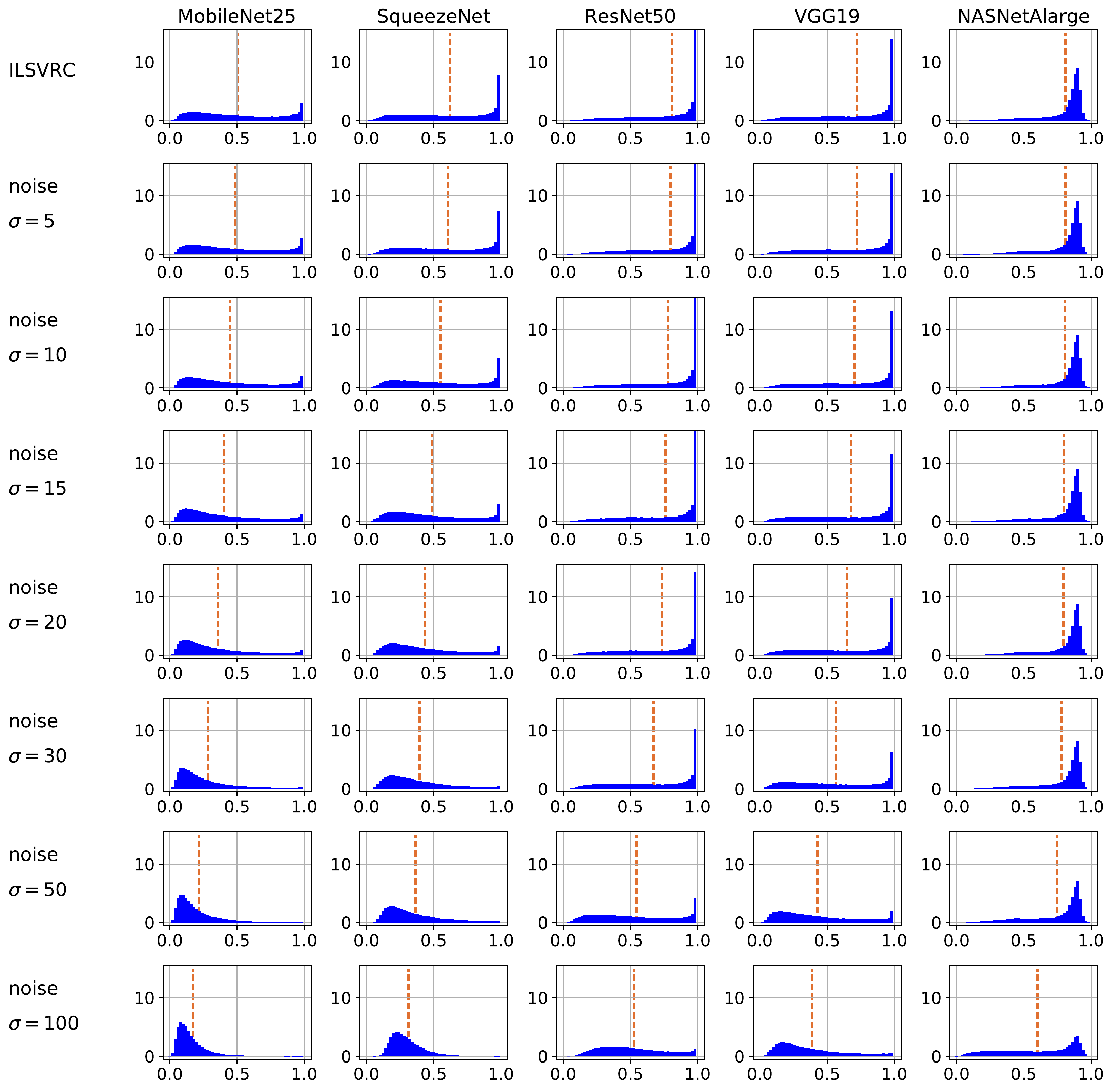}
\caption{Confidence scores for sensor noise (additive Gaussian noise) of different strengths. The dashed line marks the mean confidence.}\label{fig:distribution-noise}
\end{figure}

Figure~\ref{fig:batchsizes-noise} shows the 
detection rate achieved by \METHOD{} for different batch sizes.
Figure~\ref{fig:distribution-noise} shows the resulting distribution
of confidence scores. 
One can see that the confidence scores of all tested ConvNets are 
rather robust against additive noise, especially \emph{VGG19}
and \emph{NASNetAlarge} for which noise of strength $\sigma=5$ 
was not detectable, even for large batch sizes. For NASNetAlarge,
even $\sigma=10$ was almost undetectable from the output confidence
scores.
With increasing amounts of noise, however, the characteristics 
peaks at high confidence values start to disappear. At this level, reliable
detection is possible for all ConvNet architectures.

\clearpage
\subsubsection{Pixel Defects}
We analyze the effect of \emph{cold} and \emph{hot} dead pixel defects
by creating 100.000 perturbed test images with salt-and-pepper 
noise. We distort a random subset of $p$ percent of the pixels, 
setting half of them to pure black and half of them to pure white, 
for $p\in\{1\%,5\%,10\%,20\%,40\%,60\%,80\%,100\%\}$.

$$\tilde X_t[h,w,c] = 
\begin{cases} 0 & \text{if } (h,w)\in J_{\text{dead}}, \\
 255 & \text{if } (h,w)\in J_{\text{hot}}, \\
 X_t[h,w,c] & \text{otherwise.} \end{cases}
 \qquad\text{for $t=1,\dots,100.000$},$$
where $J_{\text{dead}}$ and $J_{\text{hot}}$ are disjoint random 
subsets of size $\lfloor\frac{1}{2}pN\rfloor$ each, where $N$ 
is the number of pixels in the image.

\begin{figure}
\centering
\adjincludegraphics[height=.3\textwidth,trim={0 0 {.47\width} 0},clip]{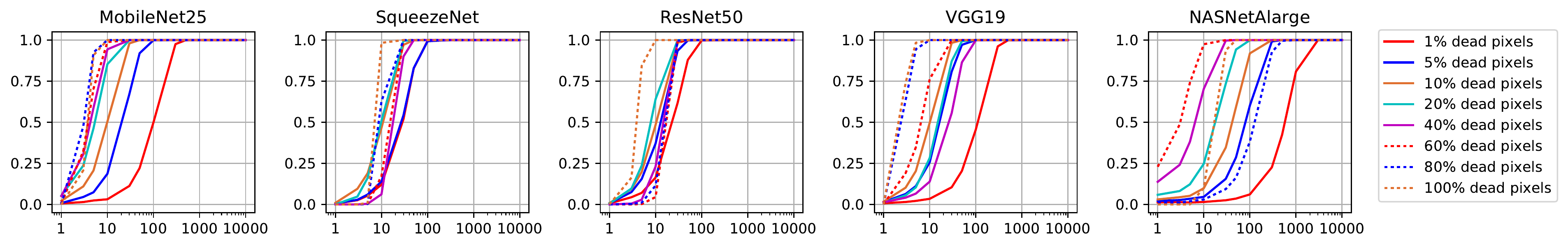}

\adjincludegraphics[height=.3\textwidth,trim={{.54\width} 0 0 0},clip]{batchsizes-dead}

\caption{Detection rates vs. \METHOD{} batch size for pixel errors (salt-and-pepper noise) of different strengths}\label{fig:batchsizes-dead}
\end{figure}

\begin{figure}\centering
\includegraphics[width=\textwidth]{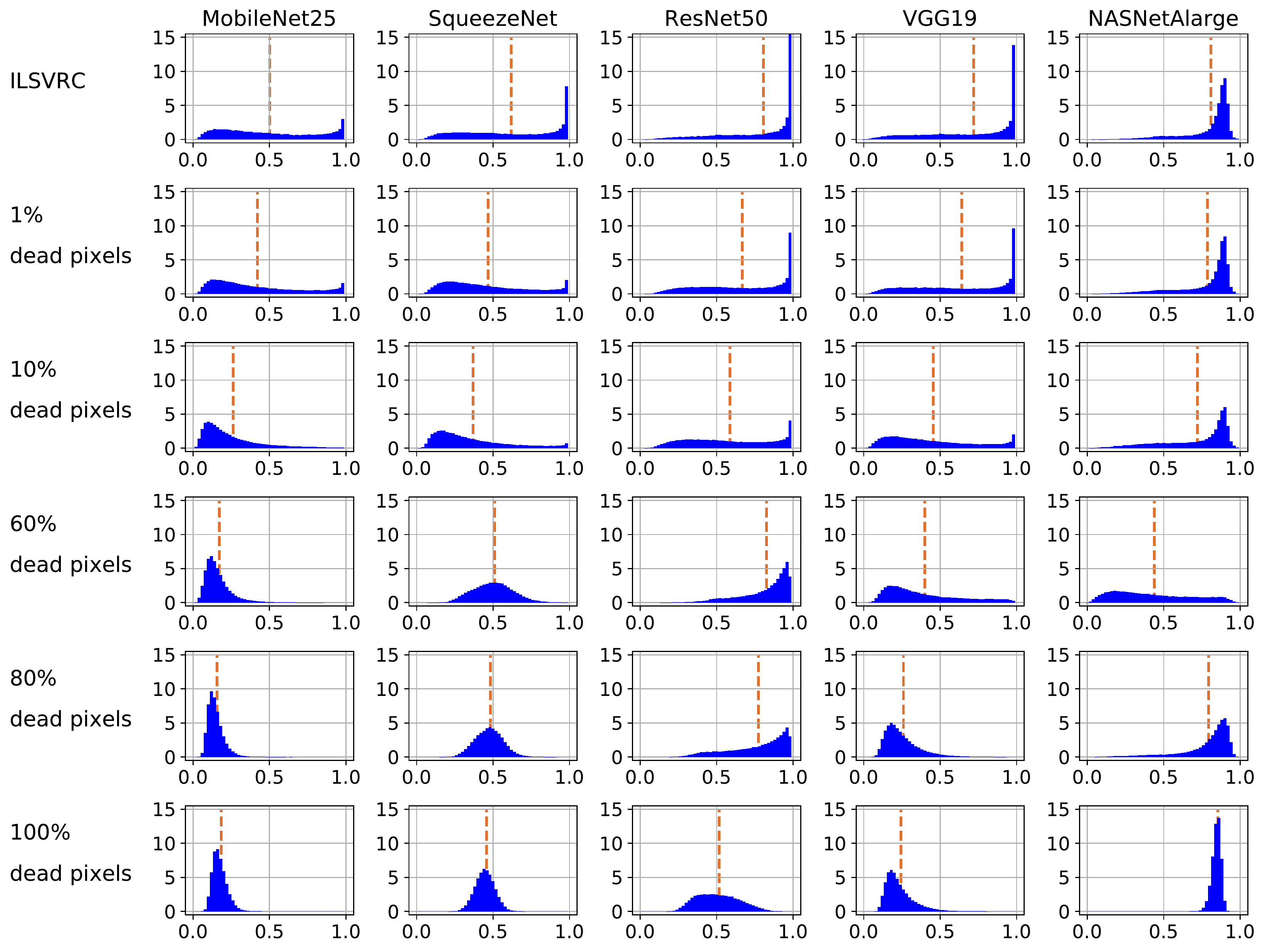}
\caption{Confidence scores for pixel errors (salt-and-pepper noise) of different strengths. The dashed line marks the mean confidence.}\label{fig:distribution-dead}
\end{figure}

Figure~\ref{fig:batchsizes-dead} shows the detection rate 
achieved by \METHOD{} for different batch sizes.
Figure~\ref{fig:distribution-dead} shows the resulting distribution
of confidence scores. 
One can see that most ConvNets are affected already by small
levels of pixel defects (\eg $p=1\%$). As for the previous 
cases, the main effect is that the distribution's peak at high 
values gets reduced, until it disappears completely.

NASNetAlarge shows a very interesting and irregular behavior,
though. 
For small perturbations ($p=1\%$), the distribution is almost
unaffected. 
When the perturbation strengths increases, until $p=60\%$ 
the expected effect happens, as the network outputs 
get overall less confident.
For even higher perturbation strength, however, NASNetAlarge gets 
more confident again. For $p=100\%$, \ie an image consisting 
purely of black and white pixels in random arrangement, NASNetAlarge 
is overall more confidence than for the undisturbed images (it 
always predicts the class label \texttt{window screen}). 
We find this an important fact to keep in mind for real-world 
applications: one of the best existing ConvNets for image 
classification makes confident predictions (of wrong labels)
when given random noise as input.

\clearpage
\subsubsection{Under- and Over-Exposure}
For each factor $c\in\{1/2,1/3,1/4,1/5,1/10,1/20,1/50,1/100\}$ 
we create 100.000 perturbed test images by scaling the intensity 
values towards 0 (under-exposure),

\begin{align*}
\tilde X_t[h,w,c] &= c \cdot X_t[h,w,c] \,\qquad\qquad\qquad\qquad\text{(under-exposure)}
\intertext{or towards 255 (over-exposure),}
\tilde X_t[h,w,c] &= 255-c\cdot(255-X_t[h,w,c]) \qquad\text{(over-exposure),}
\end{align*}
for $t=1,\dots,100.000$.

\begin{figure}
\centering
\adjincludegraphics[height=.3\textwidth,trim={0 0 {.47\width} 0},clip]{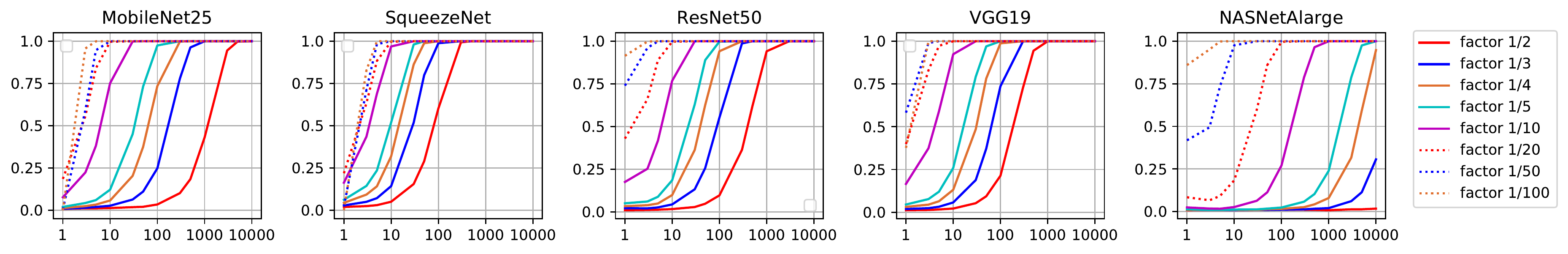}

\adjincludegraphics[height=.3\textwidth,trim={{.54\width} 0 0 0},clip]{batchsizes-dark}

\caption{Detection rates vs. \METHOD{} batch size for under-exposure (intensities scaled towards $255$ by different factors)}\label{fig:batchsizes-dark}
\end{figure}

\begin{figure}
\centering
\includegraphics[width=\textwidth]{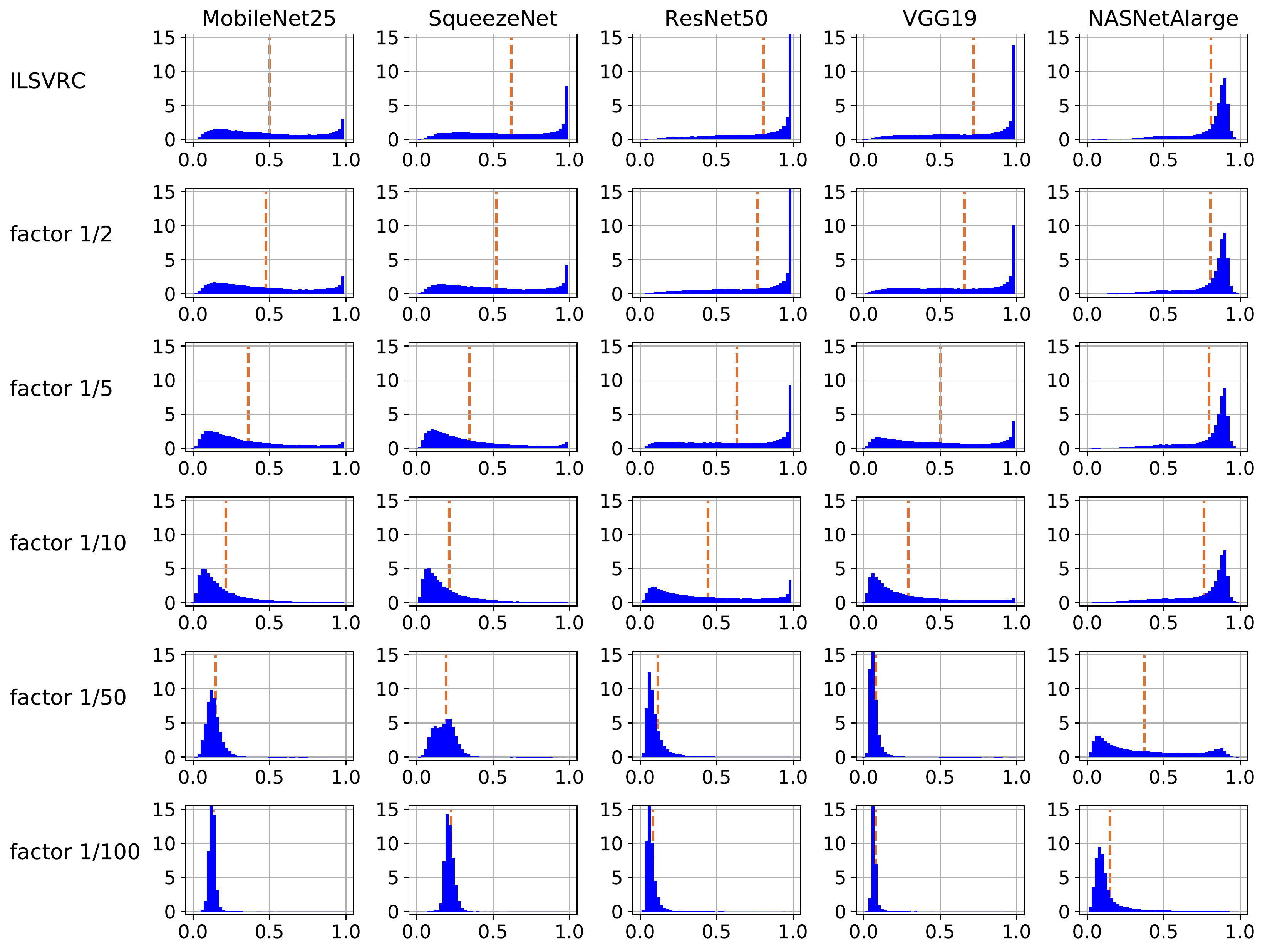}
\caption{Confidence scores for under-exposure (intensities scaled towards $0$ by different factors). The dashed line marks the mean confidence.}\label{fig:distribution-dark}
\end{figure}

\begin{figure}\centering
\adjincludegraphics[height=.3\textwidth,trim={0 0 {.47\width} 0},clip]{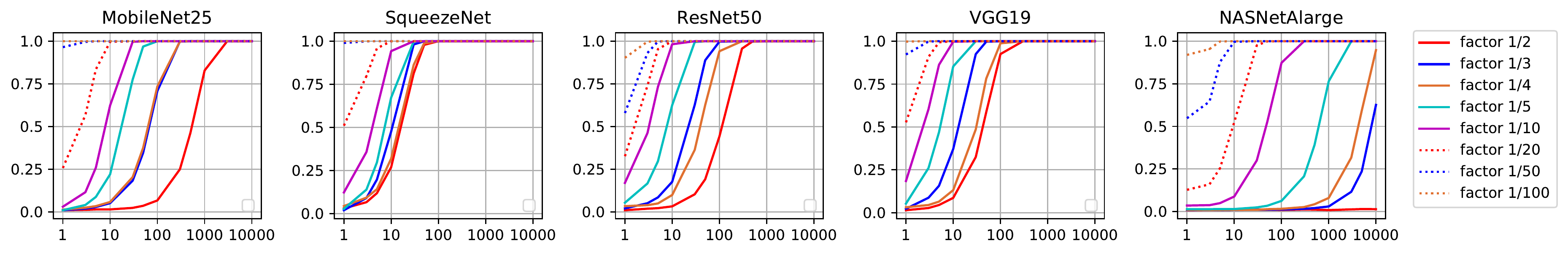}

\adjincludegraphics[height=.3\textwidth,trim={{.54\width} 0 0 0},clip]{batchsizes-bright}

\caption{Detection rates vs. \METHOD{} batch size for over-exposure (intensities scaled towards $255$ by different factors)}\label{fig:batchsizes-bright}
\end{figure}

\begin{figure}
\centering
\includegraphics[width=\textwidth]{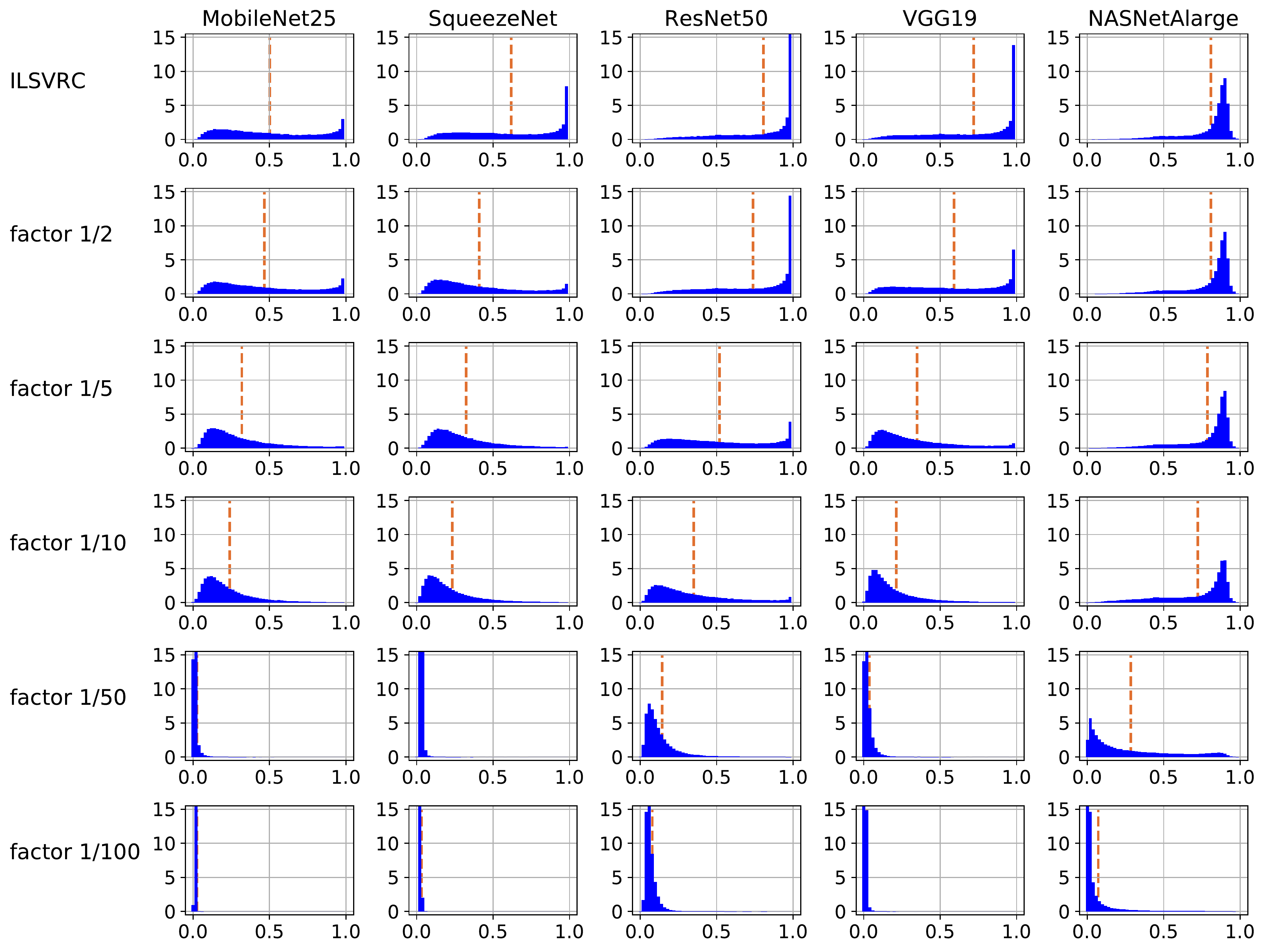}
\caption{Confidence scores for over-exposure (intensities scaled towards $0$ by different factors). The dashed line marks the mean confidence.}\label{fig:distribution-bright}
\end{figure}

Figures~\ref{fig:batchsizes-dark} and~\ref{fig:batchsizes-bright} 
show the detection rate achieved by \METHOD{} for different batch sizes.
Figures~\ref{fig:distribution-dark} and ~\ref{fig:distribution-bright} 
show the resulting distribution of confidence scores. 
One can see that the output confidences of all tested ConvNets, 
except for NASNetAlarge, are affected already by small levels 
under- or over-exposure (factor $1/2$). They react in the typical
way of predicting high confidences values less frequently. 
NASNetAlarge tolerates a much larger amount of exposure change.
A factor of $1/2$ is not detectable in the outputs, and a 
factor of $1/3$ only rarely. With bigger scale factors, its 
decisions become less confident as well, and the difference
in confidence distributions can be detected reliably,
as is the case for all other networks for all factors.

\clearpage

\FloatBarrier

\subsubsection{Geometry and Color Preprocessing}
To simulate incorrect camera installations or geometry 
preprocessing, we benchmark \METHOD{} with horizontally 
and vertically flipped images, as well as images that 
were rotated by 90, 180 or 270 degrees.
To simulate incorrect color preprocessing, we use
images in which the \texttt{R} and \texttt{B} channel
have been swapped. 
For each transformation, we create 100.000 perturbed 
test images and measure \METHOD{}'s detection rate. 

\begin{figure}\centering
\adjincludegraphics[height=.3\textwidth,trim={0 0 {.47\width} 0},clip]{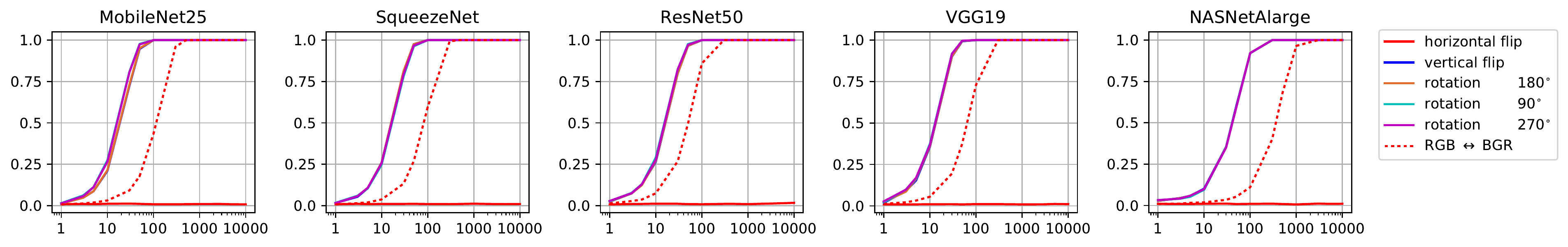}

\adjincludegraphics[height=.3\textwidth,trim={{.54\width} 0 0 0},clip]{batchsizes-flip}

\caption{Detection rates for geometric and color transformations}\label{fig:batchsizes-flip}
\end{figure}

\begin{figure}\centering
\includegraphics[width=\textwidth]{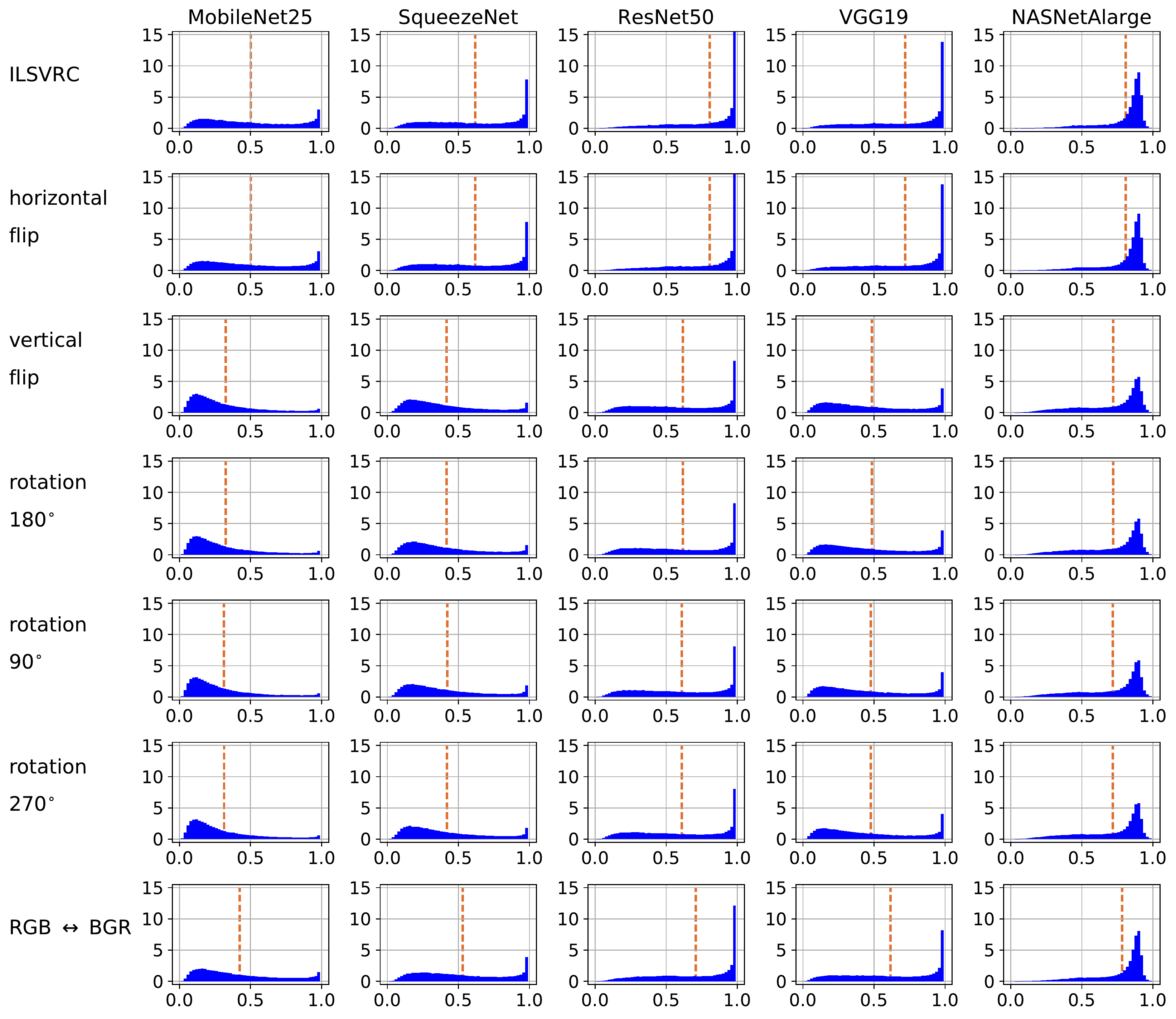}
\caption{Distribution of confidence scores for geometric and color transformations}\label{fig:distribution-flip}
\end{figure}

Figure~\ref{fig:batchsizes-flip} shows the detection 
rate achieved by \METHOD{} for different batch sizes.
Figure~\ref{fig:distribution-flip} shows the resulting 
distribution of confidence scores. 
One can see that all tested ConvNets are invariant to 
horizontal flips of the image, probably as an artifact 
of data augmentation steps that were applied during 
training. 
This could be beneficial at times, \eg when the camera 
setup requires capturing images via a mirror. It could
also be problematic in other situations, though, as it
might make it hard to fine-tuning the networks for data 
that needs left-right asymmetry, such as detecting if a 
person is left- or right-handed. 
It should also be noted that it is not guaranteed 
that every available pretrained network exhibits 
near perfect invariance to horizontal flips. 
Therefore, when relying on it for a specific task, 
it should first be confirmed, \eg using \METHOD{}. 

Vertical flips and rotations affect the confidence 
distribution of all tested ConvNets, making these 
perturbations easily detectable. 
Similarly, wrong color preprocessing is also detectable 
from the output confidences at moderate batch sizes.

\section{Conclusions}
In this work, we discussed the importance of an image 
classifier being able to identify if it is running outside 
of its specifications, \ie when the distribution of data it has 
to classify differs from the distribution of data it was trained 
for. 
We described a procedure, named \METHOD{}, based the classical 
statistical Kolmogorov-Smirnov test, which we apply to the 
distribution of the confidence values of the predicted labels. 
By extensive experiments we showed that \METHOD{} reliably 
identifies out-of-specs behavior in a variety of settings, 
including images of unexpected classes in the data, but also
effects of incorrect camera settings or sensor fatigue. 
We hope that our work leads to more research on how making 
ConvNet classifiers more trustworthy. 

\clearpage
\nocite{Tange2011a}
\bibliographystyle{spmpsci}
\bibliography{shift}

\begin{thebibliography}{10}
\providecommand{\url}[1]{{#1}}
\providecommand{\urlprefix}{URL }
\expandafter\ifx\csname urlstyle\endcsname\relax
  \providecommand{\doi}[1]{DOI~\discretionary{}{}{}#1}\else
  \providecommand{\doi}{DOI~\discretionary{}{}{}\begingroup
  \urlstyle{rm}\Url}\fi

\bibitem{ben2010theory}
Ben-David, S., Blitzer, J., Crammer, K., Kulesza, A., Pereira, F., Vaughan,
  J.W.: A theory of learning from different domains.
\newblock Machine Learning \textbf{79}(1-2), 151--175 (2010)

\bibitem{Bendale_2015_CVPR}
Bendale, A., Boult, T.: Towards open world recognition.
\newblock In: Conference on Computer Vision and Pattern Recognition (CVPR)
  (2015)

\bibitem{dunning2014computing}
Dunning, T., Ertl, O.: Computing extremely accurate quantiles using t-digests.
\newblock github.com  (2014)

\bibitem{ganin15}
Ganin, Y., Lempitsky, V.: Unsupervised domain adaptation by backpropagation.
\newblock In: International Conference on Machine Learing (ICML) (2015)

\bibitem{harel2014concept}
Harel, M., Mannor, S., El-Yaniv, R., Crammer, K.: Concept drift detection
  through resampling.
\newblock In: International Conference on Machine Learing (ICML) (2014)

\bibitem{he2016deep}
He, K., Zhang, X., Ren, S., Sun, J.: Deep residual learning for image
  recognition.
\newblock In: Conference on Computer Vision and Pattern Recognition (CVPR)
  (2016)

\bibitem{howard2017mobilenets}
Howard, A.G., Zhu, M., Chen, B., Kalenichenko, D., Wang, W., Weyand, T.,
  Andreetto, M., Adam, H.: Mobilenets: Efficient convolutional neural networks
  for mobile vision applications.
\newblock arXiv preprint arXiv:1704.04861  (2017)

\bibitem{iandola2016squeezenet}
Iandola, F.N., Han, S., Moskewicz, M.W., Ashraf, K., Dally, W.J., Keutzer, K.:
  {SqueezeNet: AlexNet-level accuracy with 50x fewer parameters and $<$0.5 MB
  model size}.
\newblock arXiv preprint arXiv:1602.07360  (2016)

\bibitem{Jain2014}
Jain, L.P., Scheirer, W.J., Boult, T.E.: Multi-class open set recognition using
  probability of inclusion.
\newblock In: European Conference on Computer Vision (ECCV) (2014)

\bibitem{kuncheva2014pca}
Kuncheva, L.I., Faithfull, W.J.: {PCA} feature extraction for change detection
  in multidimensional unlabeled data.
\newblock IEEE Transactions on Neural Networks (T-NN) \textbf{25}(1), 69--80
  (2014)

\bibitem{JSSv008i18}
Marsaglia, G., Tsang, W.W., Wang, J.: {Evaluating Kolmogorov's Distribution}.
\newblock Journal of Statistical Software, Articles \textbf{8}(18) (2003)

\bibitem{massey1951kolmogorov}
Massey~Jr, F.J.: {The Kolmogorov-Smirnov test for goodness of fit}.
\newblock Journal of the American Statistical Association \textbf{46}(253),
  68--78 (1951)

\bibitem{Perazzi2016}
Perazzi, F., Pont-Tuset, J., McWilliams, B., {Van Gool}, L., Gross, M.,
  Sorkine-Hornung, A.: A benchmark dataset and evaluation methodology for video
  object segmentation.
\newblock In: Conference on Computer Vision and Pattern Recognition (CVPR)
  (2016)

\bibitem{icarl}
Rebuffi, S.A., Kolesnikov, A., Sperl, G., Lampert, C.H.: {iCaRL: Incremental
  Classifier and Representation Learning}.
\newblock In: Conference on Computer Vision and Pattern Recognition (CVPR)
  (2016)

\bibitem{dos2016fast}
dos Reis, D.M., Flach, P., Matwin, S., Batista, G.: {Fast unsupervised online
  drift detection using incremental Kolmogorov-Smirnov test}.
\newblock In: SIGKDD (2016)

\bibitem{royer-cvpr2015}
Royer, A., Lampert, C.H.: Classifier adaptation at prediction time.
\newblock In: Conference on Computer Vision and Pattern Recognition (CVPR)
  (2015)

\bibitem{ILSVRC15}
Russakovsky, O., Deng, J., Su, H., Krause, J., Satheesh, S., Ma, S., Huang, Z.,
  Karpathy, A., Khosla, A., Bernstein, M., Berg, A.C., Fei-Fei, L.: {ImageNet
  Large Scale Visual Recognition Challenge}.
\newblock International Journal of Computer Vision (IJCV) \textbf{115}(3)
  (2015).
\newblock \doi{10.1007/s11263-015-0816-y}

\bibitem{sethi2016grid}
Sethi, T.S., Kantardzic, M., Hu, H.: A grid density based framework for
  classifying streaming data in the presence of concept drift.
\newblock Journal of Intelligent Information Systems \textbf{46}(1), 179--211
  (2016)

\bibitem{Simonyan14c}
Simonyan, K., Zisserman, A.: Very deep convolutional networks for large-scale
  image recognition.
\newblock arXiv preprint arXiv:1409.1556  (2014)

\bibitem{Tange2011a}
Tange, O.: Gnu parallel - the command-line power tool.
\newblock ;login: The USENIX Magazine \textbf{36}(1), 42--47 (2011).
\newblock \urlprefix\url{http://www.gnu.org/s/parallel}

\bibitem{wang2015concept}
Wang, H., Abraham, Z.: Concept drift detection for streaming data.
\newblock In: International Joint Conference on Neural Networks (IJCNN) (2015)

\bibitem{xian2017zero}
Xian, Y., Lampert, C.H., Schiele, B., Akata, Z.: Zero-shot learning-a
  comprehensive evaluation of the good, the bad and the ugly.
\newblock arXiv preprint arXiv:1707.00600  (2017)

\bibitem{5693384}
Zliobaite, I.: Change with delayed labeling: When is it detectable?
\newblock In: International Conference on Data Mining Workshops (2010)

\bibitem{zoph2017learning}
Zoph, B., Vasudevan, V., Shlens, J., Le, Q.V.: Learning transferable
  architectures for scalable image recognition.
\newblock arXiv preprint arXiv:1707.07012  (2017)

\end{thebibliography}

\clearpage
\section{Appendix: Tabulated Thresholds for Kolmogorov-Smirnov Test}
\label{app:thresholds}

The following values were used as thresholds for the 
Kolmogorov-Smirnov test that is part of \METHOD{}. 
They were computed using the routines provided in~\cite{JSSv008i18}.

\npdecimalsign{.}
\nprounddigits{8}
\npthousandthpartsep{}

\begin{table}[h]\centering\small
\begin{tabular}{l|n{2}{8}|n{2}{8}|n{2}{8}|n{2}{8}|n{2}{8}}
\multicolumn{1}{c|}{$\alpha$}&
\multicolumn{1}{r|}{$m=$\ \ \ \ 1}&
\multicolumn{1}{r|}{3}&
\multicolumn{1}{r|}{5}&
\multicolumn{1}{r|}{10}&
\multicolumn{1}{r}{30}\\\hline
0.00001 &0.99999500066 & 0.982894897461 & 0.912933349609 & 0.717041015625 & 0.436859130859  \\
0.00005 &0.999974995852 & 0.970764160156 & 0.8798828125 & 0.674682617188 & 0.407958984375  \\
0.0001 &0.999949991703 & 0.963165283203 & 0.861999511719 & 0.65478515625 & 0.394775390625  \\
0.0005 &0.999750137329 & 0.93701171875 & 0.809631347656 & 0.604309082031 & 0.361938476562  \\
0.001 &0.999500274658 & 0.920654296875 & 0.781372070312 & 0.580444335938 & 0.346740722656  \\
0.005 &0.997497558594 & 0.8642578125 & 0.705444335938 & 0.518737792969 & 0.308166503906  \\
0.01 &0.994995117188 & 0.828979492188 & 0.668579101562 & 0.488891601562 & 0.289855957031  \\
0.05 &0.974975585938 & 0.70751953125 & 0.563232421875 & 0.409240722656 & 0.24169921875  \\
0.1 &0.949951171875 & 0.635986328125 & 0.509521484375 & 0.36865234375 & 0.217529296875  \\
0.5 &0.75 & 0.4345703125 & 0.341796875 & 0.246826171875 & 0.145874023438  
\end{tabular}\vskip\baselineskip

\begin{tabular}{l|n{2}{8}|n{2}{8}|n{2}{8}|n{2}{8}}
\multicolumn{1}{c|}{$\alpha$}&
\multicolumn{1}{r|}{$m=$\ \ \ 50}&
\multicolumn{1}{r|}{100}&
\multicolumn{1}{r|}{300}&
\multicolumn{1}{r|}{500}\\\hline
0.00001 & 0.342071533203 & 0.243988037109 & 0.141807556152 & 0.110023498535  \\
0.00005 & 0.319091796875 & 0.227416992188 & 0.132125854492 & 0.102508544922  \\
0.0001 & 0.308624267578 & 0.219879150391 & 0.127731323242 & 0.099098205566  \\
0.0005 & 0.282653808594 & 0.201263427734 & 0.116882324219 & 0.090675354004  \\
0.001 & 0.270690917969 & 0.192687988281 & 0.111877441406 & 0.086791992188  \\
0.005 & 0.240386962891 & 0.171051025391 & 0.099304199219 & 0.077041625977 \\
0.01 & 0.226043701172 & 0.160797119141 & 0.093353271484 & 0.07243347168  \\
0.05 & 0.188415527344 & 0.134033203125 & 0.077835083008 & 0.060394287109  \\
0.1 & 0.169616699219 & 0.120666503906 & 0.070098876953 & 0.054397583008  \\
0.5 & 0.113891601562 & 0.081176757812 & 0.047241210938 & 0.036682128906  
\end{tabular}\vskip\baselineskip

\begin{tabular}{l|n{2}{8}|n{2}{8}|n{2}{8}|n{2}{8}}
\multicolumn{1}{c|}{$\alpha$}&
\multicolumn{1}{r|}{$m=1000$}&
\multicolumn{1}{r|}{3000}&
\multicolumn{1}{r|}{5000}&
\multicolumn{1}{r}{10000}\\\hline
0.00001 & 0.077911376953 & 0.045040130615 & 0.034900665283 & 0.024686813354 \\
0.00005 & 0.07258605957 & 0.041961669922 & 0.032516479492 & 0.023000717163 \\
0.0001 & 0.070171356201 & 0.040565490723 & 0.031433105469 & 0.022235870361 \\
0.0005 & 0.064208984375 & 0.037120819092 & 0.028762817383 & 0.020347595215 \\
0.001 & 0.061462402344 & 0.035533905029 & 0.027534484863 & 0.019477844238 \\
0.005 & 0.054557800293 & 0.031539916992 & 0.024444580078 & 0.017292022705 \\
0.01 & 0.051292419434 & 0.02965927124 & 0.022983551025 & 0.016258239746 \\
0.05 & 0.042778015137 & 0.024742126465 & 0.019172668457 & 0.013565063477 \\
0.1 & 0.038528442383 & 0.022285461426 & 0.017272949219 & 0.012222290039 \\
0.5 & 0.026000976562 & 0.01505279541 & 0.011672973633 & 0.00825881958 
\end{tabular}
\npnoround
\caption{Tabulated thresholds $\theta_{\alpha,m}$ for Kolmogorov-Smirnov test}
\end{table} 

\clearpage

\section{Appendix: True Positive Rates for Different ConvNets}\label{app:TPR}
\begin{table}[h]\renewcommand{\arraystretch}{.96}
\npdecimalsign{.}
\nprounddigits{2}
\npthousandthpartsep{}
\centering\scriptsize
\begin{tabular}{|l|n{2}{2}|n{2}{2}|n{2}{2}|n{2}{2}|n{2}{2}|n{2}{2}|n{2}{2}|n{2}{2}|n{2}{2}|}\hline
MobileNet25 & 
\multicolumn{1}{c|}{\rotatebox{90}{\METHOD}} &
\multicolumn{1}{c|}{\rotatebox{90}{mean}} &
\multicolumn{1}{c|}{\rotatebox{90}{$\log$-mean}} &
\multicolumn{1}{c|}{\rotatebox{90}{$z$}} &
\multicolumn{1}{c|}{\rotatebox{90}{$\log$-z}} &
\multicolumn{1}{c|}{\rotatebox{90}{sym.mean}} &
\multicolumn{1}{c|}{\rotatebox{90}{sym.$\log$-mean}} &
\multicolumn{1}{c|}{\rotatebox{90}{sym.$z$}} &
\multicolumn{1}{c|}{\rotatebox{90}{sym.$\log$-z} }
\\\hline
AwA2-bat 	& 1.0000 & 1.0000 & 1.0000 & 1.0000 & 1.0000 & 1.0000 & 1.0000 & 1.0000 & 1.0000 \\
AwA2-blue whale	& 1.0000 & 0.0000 & 0.0000 & 0.0000 & 0.0000 & 1.0000 & 0.0000 & 1.0000 & 1.0000 \\
AwA2-bobcat 	& 1.0000 & 0.0000 & 0.0000 & 0.0000 & 0.0000 & 1.0000 & 0.0000 & 1.0000 & 1.0000 \\
AwA2-dolphin	& 1.0000 & 0.9987 & 0.0000 & 0.9996 & 0.0000 & 0.9704 & 0.0000 & 0.9764 & 0.0000 \\
AwA2-giraffe	& 1.0000 & 1.0000 & 1.0000 & 1.0000 & 1.0000 & 1.0000 & 1.0000 & 1.0000 & 1.0000 \\
AwA2-horse 	& 1.0000 & 1.0000 & 1.0000 & 1.0000 & 1.0000 & 1.0000 & 1.0000 & 1.0000 & 0.9999 \\
AwA2-rat 	& 1.0000 & 1.0000 & 1.0000 & 1.0000 & 1.0000 & 1.0000 & 1.0000 & 1.0000 & 1.0000 \\
AwA2-seal 	& 1.0000 & 0.0000 & 0.0000 & 0.0000 & 0.0000 & 1.0000 & 0.0000 & 1.0000 & 1.0000 \\
AwA2-sheep	& 1.0000 & 0.0270 & 0.0000 & 0.0158 & 0.0000 & 0.0066 & 0.0000 & 0.0037 & 0.2260 \\
AwA2-walrus	& 1.0000 & 1.0000 & 1.0000 & 1.0000 & 1.0000 & 1.0000 & 1.0000 & 1.0000 & 1.0000 \\
\end{tabular}
\begin{tabular}{|l|n{2}{2}|n{2}{2}|n{2}{2}|n{2}{2}|n{2}{2}|n{2}{2}|n{2}{2}|n{2}{2}|n{2}{2}|}\hline
SqueezeNet 
\\\hline
AwA2-bat 	& 1.0000 & 1.0000 & 1.0000 & 1.0000 & 1.0000 & 1.0000 & 1.0000 & 1.0000 & 1.0000 \\
AwA2-blue whale	& 1.0000 & 0.0000 & 0.0000 & 0.0000 & 0.0000 & 1.0000 & 0.0000 & 1.0000 & 1.0000 \\
AwA2-bobcat 	& 1.0000 & 0.0000 & 0.0000 & 0.0000 & 0.0000 & 1.0000 & 0.0000 & 1.0000 & 1.0000 \\
AwA2-dolphin	& 1.0000 & 1.0000 & 0.8784 & 1.0000 & 0.9285 & 1.0000 & 0.7588 & 1.0000 & 0.5217 \\
AwA2-giraffe	& 1.0000 & 1.0000 & 1.0000 & 1.0000 & 1.0000 & 1.0000 & 1.0000 & 1.0000 & 1.0000 \\
AwA2-horse 	& 1.0000 & 1.0000 & 0.9999 & 1.0000 & 0.9998 & 1.0000 & 0.9993 & 1.0000 & 0.9993 \\
AwA2-rat 	& 1.0000 & 1.0000 & 1.0000 & 1.0000 & 1.0000 & 1.0000 & 1.0000 & 1.0000 & 1.0000 \\
AwA2-seal 	& 1.0000 & 1.0000 & 1.0000 & 1.0000 & 1.0000 & 1.0000 & 1.0000 & 1.0000 & 1.0000 \\
AwA2-sheep	& 1.0000 & 0.0000 & 0.0000 & 0.0000 & 0.0000 & 0.7651 & 0.0000 & 0.7747 & 0.9999 \\
AwA2-walrus	& 1.0000 & 1.0000 & 1.0000 & 1.0000 & 1.0000 & 1.0000 & 1.0000 & 1.0000 & 1.0000 \\
\end{tabular}
\begin{tabular}{|l|n{2}{2}|n{2}{2}|n{2}{2}|n{2}{2}|n{2}{2}|n{2}{2}|n{2}{2}|n{2}{2}|n{2}{2}|}\hline
ResNet50
\\\hline
AwA2-bat 	& 1.0000 & 1.0000 & 1.0000 & 1.0000 & 1.0000 & 1.0000 & 1.0000 & 1.0000 & 1.0000 \\
AwA2-blue whale	& 1.0000 & 1.0000 & 0.0000 & 1.0000 & 0.0000 & 0.9995 & 0.0000 & 1.0000 & 0.0000 \\
AwA2-bobcat 	& 1.0000 & 0.0000 & 0.0000 & 0.0000 & 0.0000 & 1.0000 & 0.0000 & 1.0000 & 1.0000 \\
AwA2-dolphin	& 1.0000 & 1.0000 & 1.0000 & 1.0000 & 1.0000 & 1.0000 & 1.0000 & 1.0000 & 1.0000 \\
AwA2-giraffe	& 1.0000 & 1.0000 & 1.0000 & 1.0000 & 1.0000 & 1.0000 & 1.0000 & 1.0000 & 1.0000 \\
AwA2-horse 	& 1.0000 & 1.0000 & 1.0000 & 1.0000 & 1.0000 & 1.0000 & 1.0000 & 1.0000 & 1.0000 \\
AwA2-rat 	& 1.0000 & 1.0000 & 1.0000 & 1.0000 & 1.0000 & 1.0000 & 1.0000 & 1.0000 & 1.0000 \\
AwA2-seal 	& 1.0000 & 0.0000 & 0.0000 & 0.0000 & 0.0000 & 1.0000 & 0.0000 & 1.0000 & 1.0000 \\
AwA2-sheep	& 1.0000 & 0.0000 & 0.0000 & 0.0000 & 0.0000 & 0.4818 & 0.0000 & 0.3755 & 0.8472 \\
AwA2-walrus	& 1.0000 & 1.0000 & 1.0000 & 1.0000 & 1.0000 & 1.0000 & 1.0000 & 1.0000 & 1.0000 \\
\end{tabular}
\begin{tabular}{|l|n{2}{2}|n{2}{2}|n{2}{2}|n{2}{2}|n{2}{2}|n{2}{2}|n{2}{2}|n{2}{2}|n{2}{2}|}\hline
VGG19 
\\\hline
AwA2-bat 	& 1.0000 & 1.0000 & 1.0000 & 1.0000 & 1.0000 & 1.0000 & 1.0000 & 1.0000 & 1.0000 \\
AwA2-blue whale	& 1.0000 & 0.0000 & 0.0000 & 0.0000 & 0.0000 & 0.0000 & 0.0000 & 0.0000 & 0.0797 \\
AwA2-bobcat 	& 1.0000 & 0.0000 & 0.0000 & 0.0000 & 0.0000 & 1.0000 & 0.0000 & 1.0000 & 1.0000 \\
AwA2-dolphin	& 1.0000 & 1.0000 & 1.0000 & 1.0000 & 1.0000 & 1.0000 & 1.0000 & 1.0000 & 1.0000 \\
AwA2-giraffe	& 1.0000 & 1.0000 & 1.0000 & 1.0000 & 1.0000 & 1.0000 & 1.0000 & 1.0000 & 1.0000 \\
AwA2-horse 	& 1.0000 & 1.0000 & 1.0000 & 1.0000 & 1.0000 & 1.0000 & 1.0000 & 1.0000 & 1.0000 \\
AwA2-rat 	& 1.0000 & 1.0000 & 1.0000 & 1.0000 & 1.0000 & 1.0000 & 1.0000 & 1.0000 & 1.0000 \\
AwA2-seal 	& 1.0000 & 0.0000 & 0.0000 & 0.0000 & 0.0000 & 1.0000 & 0.0000 & 1.0000 & 1.0000 \\
AwA2-sheep	& 1.0000 & 0.0000 & 0.0000 & 0.0000 & 0.0000 & 0.9959 & 0.0000 & 0.9856 & 1.0000 \\
AwA2-walrus	& 1.0000 & 1.0000 & 1.0000 & 1.0000 & 1.0000 & 1.0000 & 1.0000 & 1.0000 & 1.0000 \\
\end{tabular}
\begin{tabular}{|l|n{2}{2}|n{2}{2}|n{2}{2}|n{2}{2}|n{2}{2}|n{2}{2}|n{2}{2}|n{2}{2}|n{2}{2}|}\hline
NASNetAlarge
\\\hline
AwA2-bat 	& 1.0000 & 1.0000 & 1.0000 & 1.0000 & 1.0000 & 1.0000 & 1.0000 & 1.0000 & 1.0000 \\
AwA2-blue whale	& 1.0000 & 0.0000 & 0.0000 & 0.0000 & 0.0000 & 1.0000 & 0.0000 & 1.0000 & 1.0000 \\
AwA2-bobcat 	& 1.0000 & 0.0000 & 0.0000 & 0.0000 & 0.0000 & 1.0000 & 0.0000 & 1.0000 & 1.0000 \\
AwA2-dolphin	& 1.0000 & 1.0000 & 1.0000 & 1.0000 & 1.0000 & 1.0000 & 1.0000 & 1.0000 & 1.0000 \\
AwA2-giraffe	& 1.0000 & 1.0000 & 1.0000 & 1.0000 & 1.0000 & 1.0000 & 1.0000 & 1.0000 & 1.0000 \\
AwA2-horse 	& 1.0000 & 1.0000 & 1.0000 & 1.0000 & 1.0000 & 1.0000 & 1.0000 & 1.0000 & 1.0000 \\
AwA2-rat 	& 1.0000 & 1.0000 & 1.0000 & 1.0000 & 1.0000 & 1.0000 & 1.0000 & 1.0000 & 1.0000 \\
AwA2-seal 	& 1.0000 & 0.0000 & 0.0000 & 0.0000 & 0.0000 & 1.0000 & 0.0000 & 1.0000 & 1.0000 \\
AwA2-sheep	& 1.0000 & 0.0000 & 0.0000 & 0.0000 & 0.0000 & 1.0000 & 0.0000 & 1.0000 & 1.0000 \\
AwA2-walrus	& 1.0000 & 1.0000 & 1.0000 & 1.0000 & 1.0000 & 1.0000 & 1.0000 & 1.0000 & 1.0000 \\
\hline
\end{tabular}
\npnoround
\caption{True positive rate of \METHOD{} and baselines under different out-of-specs conditions
for ConvNets \emph{MobileNet25}, \emph{SqueezeNet}, \emph{ResNet50}, \emph{VGG19}, \emph{NASNetAlarge}.}\label{tab:TPR-allnets}
\end{table}

\clearpage
\section{Appendix: Illustrations of Synthetic Image Manipulations}\label{app:synthetic}
\subsection{\emph{Loss-of-focus} (Gaussian blur)}
\begin{center}
\begin{tabular}{ccc}
\includegraphics[width=.28\textwidth]{zebra.jpg}&
\includegraphics[width=.28\textwidth]{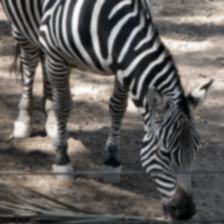}&
\includegraphics[width=.28\textwidth]{zebrablur2.png}\\
original & $\sigma=1$ & $\sigma=2$ \\
\includegraphics[width=.28\textwidth]{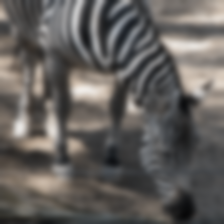}&
\includegraphics[width=.28\textwidth]{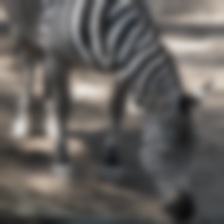}&
\includegraphics[width=.28\textwidth]{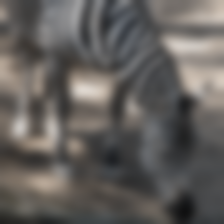}\\
$\sigma=3$  & $\sigma=4$ & $\sigma=5$\\
\includegraphics[width=.28\textwidth]{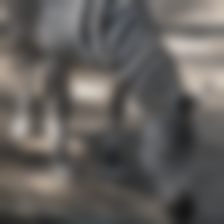}&
\includegraphics[width=.28\textwidth]{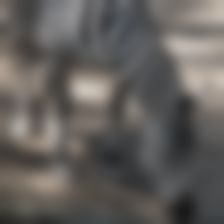}&
\includegraphics[width=.28\textwidth]{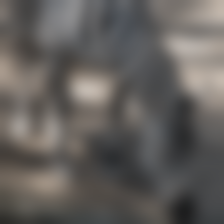}\\
$\sigma=6$ & $\sigma=7$ & $\sigma=8$ \\
\includegraphics[width=.28\textwidth]{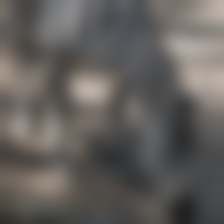}&
\includegraphics[width=.28\textwidth]{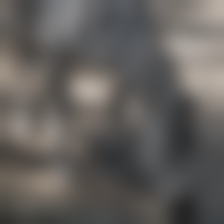}\\
$\sigma=9$ & $\sigma=10$
\end{tabular}
\end{center}

\clearpage\subsection{\emph{Sensor noise} (additive Gaussian noise)}

\begin{center}
\begin{tabular}{ccc}
\includegraphics[width=.3\textwidth]{zebra.jpg}&
\includegraphics[width=.3\textwidth]{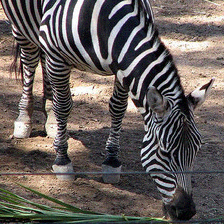}& 
\includegraphics[width=.3\textwidth]{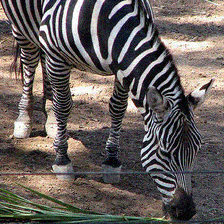} \\
original & $\sigma=5$ & $\sigma=10$ \\
\includegraphics[width=.3\textwidth]{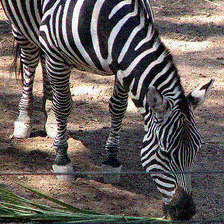} & 
\includegraphics[width=.3\textwidth]{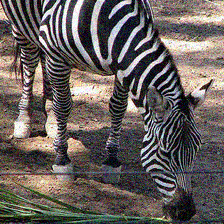} &
\includegraphics[width=.3\textwidth]{zebranoise30.png} \\
$\sigma=15$ & $\sigma=20$ & $\sigma=30$ \\
\includegraphics[width=.3\textwidth]{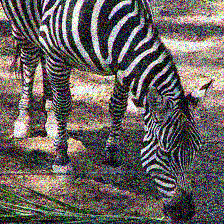} &
\includegraphics[width=.3\textwidth]{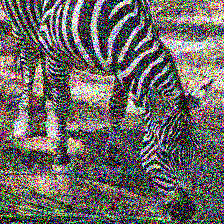} \\
$\sigma=50$ & $\sigma=100$ 
\end{tabular}
\end{center}

\clearpage\subsection{\emph{Pixel defects} (salt-and-pepper noise)}

\begin{center}
\begin{tabular}{ccc}
\includegraphics[width=.3\textwidth]{zebra.jpg}&
\includegraphics[width=.3\textwidth]{zebradead1.png} &
\includegraphics[width=.3\textwidth]{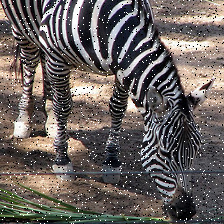} \\
original & $p=1\%$ & $p=5\%$ \\
\includegraphics[width=.3\textwidth]{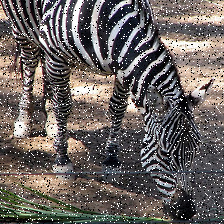} &
\includegraphics[width=.3\textwidth]{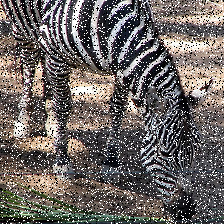} &
\includegraphics[width=.3\textwidth]{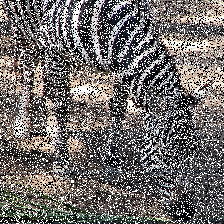} \\
$p=10\%$ & $p=20\%$ & $p=40\%$\\
\includegraphics[width=.3\textwidth]{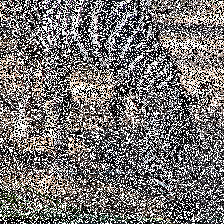} & 
\includegraphics[width=.3\textwidth]{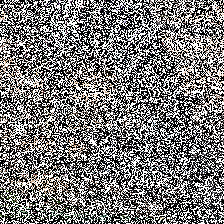}& 
\includegraphics[width=.3\textwidth]{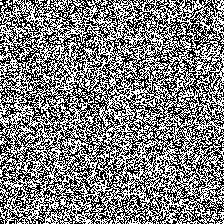}\\
$p=60\%$ & $p=80\%$ & $p=100\%$ 
\end{tabular}
\end{center}

\clearpage\subsection{\emph{Under-exposure} (scaling towards 0)}

\begin{center}
\begin{tabular}{ccc}
\includegraphics[width=.3\textwidth]{zebra.jpg}&
\includegraphics[width=.3\textwidth]{zebrascale2.png} &
\includegraphics[width=.3\textwidth]{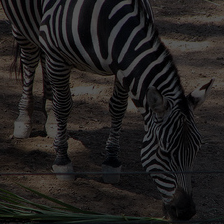} \\
original  & factor 1/2 & factor 1/3 \\
\includegraphics[width=.3\textwidth]{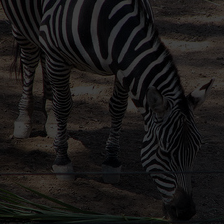} & 
\includegraphics[width=.3\textwidth]{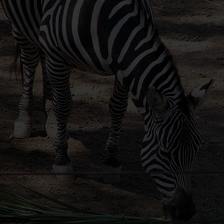}& 
\includegraphics[width=.3\textwidth]{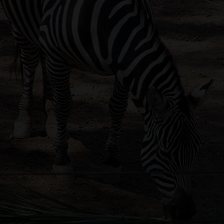}\\
factor 1/4 & factor 1/5 & factor 1/10 \\
\includegraphics[width=.3\textwidth]{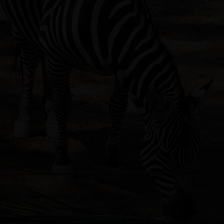} & 
\includegraphics[width=.3\textwidth]{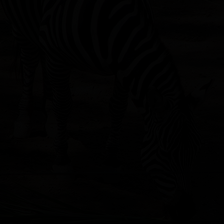} & 
\includegraphics[width=.3\textwidth]{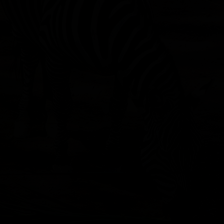}\\
factor 1/20  & factor 1/50 & factor 1/100 
\end{tabular}
\end{center}

\clearpage\subsection{\emph{Over-exposure} (scaling towards 255)}

\begin{center}
\begin{tabular}{ccc}
\includegraphics[width=.3\textwidth]{zebra.jpg}&
\includegraphics[width=.3\textwidth]{zebrascale-2.png} & 
\includegraphics[width=.3\textwidth]{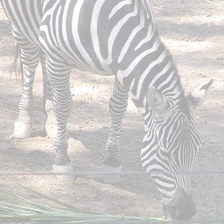} \\
original  & factor 1/2 & factor 1/3 \\
\includegraphics[width=.3\textwidth]{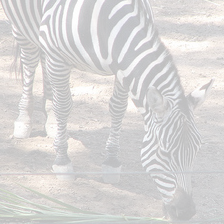} & 
\includegraphics[width=.3\textwidth]{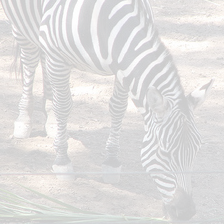} &
\includegraphics[width=.3\textwidth]{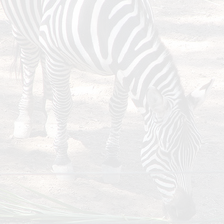} \\
factor 1/4 & factor 1/5 & factor 1/10 \\
\includegraphics[width=.3\textwidth]{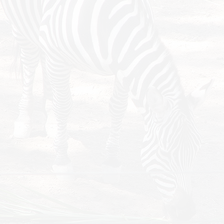} & 
\includegraphics[width=.3\textwidth]{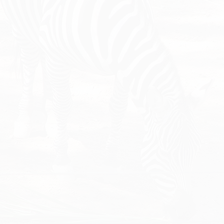}& 
\includegraphics[width=.3\textwidth]{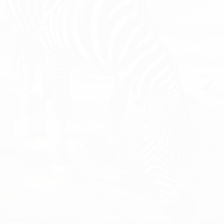}\\
factor 1/20  & factor 1/50 & factor 1/100 
\end{tabular}
\end{center}

\clearpage\subsection{\emph{Wrong geometry} and \emph{color preprocessing}}

\begin{center}
\begin{tabular}{ccc}
\includegraphics[width=.29\textwidth]{zebra.jpg}&
\includegraphics[width=.29\textwidth]{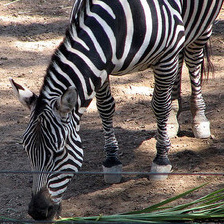}&
\includegraphics[width=.29\textwidth]{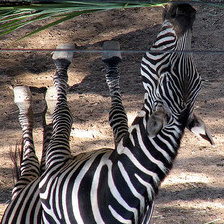}\\
original & horizontal flip & vertical flip    \\
\includegraphics[width=.29\textwidth]{zebraflip4.png}&
\includegraphics[width=.29\textwidth]{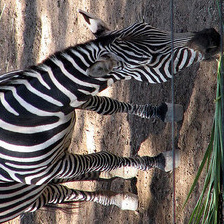}&
\includegraphics[width=.29\textwidth]{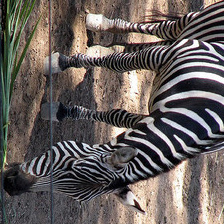}\\
180${}^\circ$ rotation & 90${}^\circ$ rotation & 270${}^\circ$ rotation \\
\includegraphics[width=.29\textwidth]{zebraBGR.png}\\
\texttt{RGB} $\longleftrightarrow$ \texttt{BGR}
\end{tabular}
\end{center}

\end{document}